\documentclass[final, 12pt]{elsarticle}
\usepackage{physics}
\usepackage[shortlabels]{enumitem}
\usepackage{mathtools}
\usepackage{amsthm}
\usepackage[utf8]{inputenc}
\usepackage{graphicx}
\usepackage{mathtools}
\usepackage[ruled,linesnumbered]{algorithm2e}
\usepackage[T1]{fontenc}
\usepackage{tabularx}
\usepackage{physics}
\usepackage{booktabs}
\graphicspath{ {images/} }
\usepackage[english]{babel}
\usepackage[margin=1in]{geometry}
\usepackage{url}
\usepackage[ruled,linesnumbered]{algorithm2e}
\usepackage{amsmath, nccmath}
\usepackage{float}
\usepackage{caption}
\usepackage{comment}
\usepackage{lscape}
\usepackage{afterpage}
\usepackage{bookmark}
\usepackage{float}
\usepackage{subcaption}
\usepackage{comment}
\usepackage{esint}
\usepackage{placeins}

\makeatletter
\def\ps@pprintTitle{%
  \let\@oddhead\@empty
  \let\@evenhead\@empty
  \let\@oddfoot\@empty
  \let\@evenfoot\@oddfoot
}
\makeatother

\newcommand{\diff}{\,\text{d}}
\newcommand{\dep}{\text{dep}}
\newcommand{\bslash}{\,\setminus\,}

\newcommand\restr[2]{{
		\left.\kern-\nulldelimiterspace 
		#1 
		\vphantom{\big|} 
		\right|_{#2} 
}}

\newcommand{\imp}{\mathcal{I}}
\newcommand{\pwset}{\mathcal{P}}

\usepackage[utf8]{inputenc}
\usepackage{lmodern}
\usepackage{mathtools}
\newtheorem{thm}{Theorem}[section] 
\newtheorem{proposition}[thm]{Proposition}

\newtheorem{cor}[thm]{Corollary}
\newtheorem{remark}[thm]{Remark}
\newtheorem{definition}[thm]{Definition} 
\newtheorem{example}[thm]{Example}
\usepackage{amssymb}
\usepackage{esint}
\usepackage{hyperref}
\begin{document}
\begin{frontmatter}

	\title{Feature Subset Weighting for Distance-based Supervised Learning through Choquet Integration}
	\author[mymainaddress]{Adnan Theerens\corref{mycorrespondingauthor}}
	\cortext[mycorrespondingauthor]{Corresponding author}
	\ead{adnan.theerens@ugent.be}
	\author[mymainaddress,secondaddress]{Yvan Saeys}
	\ead{yvan.saeys@ugent.be}
	\author[mymainaddress]{Chris Cornelis}
	\ead{chris.cornelis@ugent.be}
	\address[mymainaddress]{Department of Mathematics, Computer Science and Statistics, Ghent University, Ghent, Belgium}
	\address[secondaddress]{Data Mining and Modeling for Biomedicine Group,
VIB Inflammation Research Center}
\begin{abstract}
	This paper introduces feature subset weighting using monotone measures for distance-based supervised learning. The Choquet integral is used to define a distance metric that incorporates these weights. This integration enables the proposed distances to effectively capture non-linear relationships and account for interactions both between conditional and decision attributes and among conditional attributes themselves, resulting in a more flexible distance measure. In particular, we show how this approach ensures that the distances remain unaffected by the addition of duplicate and strongly correlated features. Another key point of this approach is that it makes feature subset weighting computationally feasible, since only \(m\) feature subset weights should be calculated each time instead of calculating all feature subset weights (\(2^m\)), where \(m\) is the number of attributes. Next, we also examine how the use of the Choquet integral for measuring similarity leads to a non-equivalent definition of distance. The relationship between distance and similarity is further explored through dual measures. Additionally, symmetric Choquet distances and similarities are proposed, preserving the classical symmetry between similarity and distance. Finally, we introduce a concrete feature subset weighting distance, evaluate its performance in a \(k\)-nearest neighbors (KNN) classification setting, and compare it against Mahalanobis distances and weighted distance methods.
	\end{abstract}


\end{frontmatter}

\section{Introduction}
In machine learning, there is a continuous effort to develop algorithms that are not only more effective and robust but also more interpretable. A core concept in many of these techniques is the measurement of similarity or dissimilarity (distances) between data instances, often achieved through distance metrics. These distances underpin many supervised learning algorithms such as \(k\)-nearest neighbours (KNN), support vector machine (SVM), and rule induction algorithms \cite{BOLLAERT2025121362}. They are also central to unsupervised learning methods like k-means clustering and density-based clustering \cite{dbscan}, as well as anomaly detection approaches such as the local outlier factor (LOF) \cite{breunig2000lof}.

Traditional distance measures, such as Euclidean and Manhattan distance, have been widely used in machine learning techniques. However, as datasets become more complex and diverse, these measures may not always capture the nuances of intricate data relationships. There is a growing interest in developing more flexible and adaptive distance metrics, particularly for tasks that require distinguishing between classes effectively. In classification, ensuring that neighbors belong to the same class improves the separation of data into distinct categories. The aim is to bring same-class instances closer together while pushing different-class instances farther apart.

Weighted distance metrics \cite{paredes2006learning,wettschereck1997review,zhang2022k} have emerged as an effective approach, assigning weights to attributes based on their relevance to the decision attribute. While this method accounts for simple relationships, it often struggles to capture more intricate patterns within the data. To address this limitation, the Mahalanobis distance incorporates correlations among attributes by applying a linear transformation to the features before calculating the distance. In metric learning \cite{bellet2015metric,nguyen2018approach}, a primary objective is to learn a Mahalanobis distance such that similar data points are brought closer together while dissimilar points are pushed further apart, based on the task-specific similarity or dissimilarity constraints. Despite its strengths, the Mahalanobis distance has a significant drawback: the transformation obscures the original attributes, reducing interpretability and making it difficult to understand the role of individual features in the distance computation.

This paper introduces a novel distance metric for supervised learning based on the Choquet integral, designed to retain the interpretability of the original attributes while capturing complex interactions.  The Choquet integral, an extension of the Lebesgue integral to non-additive measures, is widely used in decision-making contexts for its ability to model non-linear interactions \cite{grabisch2010decade}. We believe that, in the future, the Choquet distance could serve as a viable alternative to the Mahalanobis distance for metric learning.

In supervised learning, it enables the consideration of interactions among conditional attributes relative to the target attribute, offering a powerful and elegant framework for improving performance. However, the Choquet integral has only been used sporadically in the past for supervised learning or for creating distance measures. In \cite{bolton2008discrete}, the authors characterize the class of measures that induce a metric through the Choquet integral. Moreover, \cite{abril2012choquet} introduces a distance-based record linkage method that uses the Choquet integral to compute distances between records, employing a learning approach to determine the optimal non-additive measure for the linkage process. Similarly, \cite{beliakov2011learning} presents a method for learning a Choquet distance for clustering. Furthermore, \cite{ma2017choquetdistancesclassification} proposes a nonlinear classifier based on the Choquet integral with respect to a signed efficiency measure, where the decision boundary is a Choquet broken-hyperplane. In \cite{theerens2022fedcsis,THEERENS2024120385,theerens2022choquet}, the authors utilize the Choquet integral to design a noise-tolerant fuzzy rough set model, enhancing its effectiveness for classification and feature selection. Despite these contributions, none of these studies explores the potential of the Choquet integral for defining distances tailored to distance-based supervised learning algorithms.

The remainder of this paper is organized as follows: in Section \ref{sec: prelims}, we recall the required prerequisites. Section \ref{sec: FRdistances} introduces Choquet distances, provides an example, and discusses the computational complexity of calculating them. Section \ref{sec: attribute importance measures} presents a concrete attribute importance measure, based on a dependency measure from fuzzy rough set theory, to be used with Choquet distances in supervised learning. Section \ref{sec: choquet similarities} explores the duality of similarity and distance, characterizing cases where Choquet distances do not fully satisfy this duality. Section \ref{sec: Choquet distances seen as feature subset weighted distances} provides a deeper understanding of how Choquet distances can be interpreted as feature subset-weighted distances. In Section \ref{sec: duplicatesChoquet}, we examine how Choquet distances handle duplicate features. Section \ref{sec: experiments} applies and tests these novel distances for classification using KNN both on synthetic data and real-life benchmark datasets, demonstrating their potential as well as their stability under the addition of duplicate features. Finally, Section \ref{sec: conclusion} concludes the paper and outlines directions for future research.

\section{Preliminaries}
\label{sec: prelims}

\subsection{Choquet integral}
Since we will view the Choquet integral as an aggregation operator, we will restrict ourselves to measures (and Choquet integrals) on finite sets. For the general setting, we refer the reader to e.g.\ \cite{wang2010generalized}. We will use the notation \(\mathcal{P}(X)\) to represent the powerset of \(X\) throughout this paper.
\begin{definition}
	Let \(X\) be a finite set. A function \(\mu:\mathcal{P}(X)\to[0,+\infty[\) is called a \emph{monotone measure} if:
	\[\mu(\emptyset)=0 \text{ and }\: (\forall A,B\in\mathcal{P}(X))(A\subseteq B \implies \mu(A)\leq \mu(B)).\]
	A monotone measure is called \emph{additive} if \(\mu(A\cup B)=\mu(A)+\mu(B)\) when \(A\) and \(B\) are disjoint. If \(\mu(X)=1\), we call \(\mu\) normalized.
\end{definition}

\begin{definition}[\cite{wang2010generalized}]
	\label{defn: ChoquetIntegral}
	The \emph{Choquet integral} of \(f:X\to\mathbb{R}\) with respect to a monotone measure $\mu$ on \(X\) is defined as:
	\begin{equation*}
		\int f(x) \diff \mu(x)=\sum_{i=1}^n f(x^\ast_i)\cdot\left[\mu(A^\ast_i)-\mu(A^\ast_{i+1})\right],
	\end{equation*}
	where \((x^\ast_1,x^\ast_2,\dots,x^\ast_n)\) is a permutation of \(X=(x_1,x_2,\dots,x_n)\) such that
	\begin{equation*}
		f(x^\ast_1)\leq f(x^\ast_2) \leq\cdots\leq f(x^\ast_n),
	\end{equation*}
	\(A^\ast_i:=\{x^\ast_i,\dots,x^\ast_n\}\) and \(\mu(A^\ast_{n+1}):=0\). If the integration variable \(x\) is clear from the context, we use the simplified notation \(\int f \diff \mu\).
\end{definition}
Equivalently, by rearranging the terms of the sum, the Choquet integral can be defined as:
\begin{proposition}\cite{wang2010generalized}
	\label{prop: ChoquetIntegral2}
	Let $\mu$ be a monotone measure on \(X\) and \(f:X\to\mathbb{R}\) a real-valued function. The \emph{Choquet integral} of \(f\) with respect to the measure $\mu$ is equal to:
	\begin{equation*}
		\int f \diff \mu=\sum_{i=1}^{n}\mu(\{x^\ast_i,\dots,x^\ast_n\})\cdot\left[f(x^\ast_i)-f(x^\ast_{i-1})\right],
	\end{equation*}
	where \((x^\ast_1,x^\ast_2,\dots,x^\ast_n)\) is a permutation of \(X=(x_1,x_2,\dots,x_n)\) such that
	\begin{equation*}
		f(x^\ast_1)\leq f(x^\ast_2) \leq\cdots\leq f(x^\ast_n),
	\end{equation*}
	and \(f(x^\ast_0):=0\).
\end{proposition}
A characterization of the Choquet integral that is particularly useful for the implementation of the Choquet distance, is the following:
\begin{proposition}
	\label{prop: ChoquetIntegral3}
	The \emph{Choquet integral} of \(f:X\to\mathbb{R}\) with respect to a monotone measure $\mu$ on \(X\) can be written as:
	\begin{equation*}
		\int f \diff \mu=\sum_{i=1}^n \mu(B^\ast_i)\cdot\left[f(y^\ast_i)-f(y^\ast_{i+1})\right],
	\end{equation*}
	where \((y^\ast_1,y^\ast_2,\dots,y^\ast_n)\) is a permutation of \(X=(x_1,x_2,\dots,x_n)\) such that
	\begin{equation*}
		f(y^\ast_1)\geq f(y^\ast_2) \geq\cdots\geq f(y^\ast_n),
	\end{equation*}
	\(B^\ast_i:=\{y^\ast_1,\dots,y^\ast_i\}\) and \(f(y^\ast_{n+1}):=0\).
\end{proposition}
\begin{proof}
	Substituting \(j = n-i+1\) in Proposition \ref{prop: ChoquetIntegral2} gives us the desired result.
\end{proof}
One subclass of operators within the Choquet integral is the weighted sum:
\begin{proposition}{\cite{beliakov2007aggregation}}
	\label{prop: choqAdditive}
	The Choquet integral with respect to an additive measure \(\mu\) is equal to
	\[\int f \diff \mu = \sum_{x\in X} f(x)\mu(\{x\}).\]
\end{proposition}
\begin{example}
	\label{exmp: counting}
	The Choquet integral with respect to the counting measure, i.e.\ \(\mu_\#(A) = |A|\), is equal to the sum:
	\[\int f \diff \mu_\# = \sum_{x\in X} f(x).\]
\end{example}
The following propositions prove to be useful throughout the remainder of the paper.
\begin{proposition}\cite{wang2010generalized}
	\label{choqintegralDual}
	Suppose $\mu$ is a monotone measure on \(X\) and \(f:X\to\mathbb{R}\) a real-valued function, then we have the following equality:
	\[\int (-f)\diff \mu = -\int f \diff \overline{\mu},\]
	where \(\overline{\mu}\) is the dual of \(\mu\) defined by
	\[\overline{\mu}(A)= \mu(X)-\mu\left(X\bslash A\right),\;\;\forall A\subseteq X.\]
\end{proposition}

\begin{proposition}{\cite{wang2010generalized}}
	\label{prop: monotoneChoquet}
	Suppose \(\mu_1,\mu_2\) are two monotone measures with \(\mu_1 \leq \mu_2\), then for any \(f:X\to\mathbb{R}\) we have:
	\[\int f \diff \mu_1  \leq \int f \diff \mu_2.\]
\end{proposition}

The Möbius transform is a useful tool in the context of the Choquet integral. Given a monotone measure \(\mu\) on a finite set \(X\), the Möbius transform \(\mathcal{M}_\mu\) is defined as:

\[
    \mathcal{M}_\mu(B) = \sum_{A \subseteq B} (-1)^{|B| - |A|} \mu(A), \quad B \subseteq X.
\]

Using the Möbius transform, the Choquet integral can be rewritten:
\begin{proposition}\cite{beliakov2007aggregation}
	\label{mobiusChoquet}
	Suppose $\mu$ is a monotone measure on \(X\) and \(f:X\to\mathbb{R}\) a real-valued function. We can rewrite the Choquet integral in terms of the Möbius transform \(\mathcal{M}_\mu\) of the measure \(\mu\) as follows:
	\[
    \int f \, \mathrm{d} \mu = \sum_{B \subseteq X} \mathcal{M}_\mu(B) \min_{x \in B} f(x).\]
\end{proposition}
\begin{proposition}
	\label{mobiusChoquetDual}
	Suppose $\mu$ is a monotone measure on \(X\) and \(f:X\to\mathbb{R}\) a real-valued function, then we have the following:
	\[
    \int f \, \mathrm{d} \overline{\mu}= \sum_{B \subseteq X} \mathcal{M}_\mu(B) \max_{x \in B} f(x).\]
\end{proposition}
\begin{proof}
	Using Proposition \ref{choqintegralDual} and Proposition \ref{mobiusChoquet}, we have:
	\begin{align*}
		\int f \, \mathrm{d} \overline{\mu} &= - \int (-f) \diff \mu =-\sum_{B \subseteq X} \mathcal{M}_\mu(B) \min_{x \in B} (-f(x))\\
		&=-\sum_{B \subseteq X} \mathcal{M}_\mu(B) (-\max_{x \in B} f(x))=\sum_{B \subseteq X} \mathcal{M}_\mu(B) \max_{x \in B} f(x).
	\end{align*}
\end{proof}
\subsection{Distance and similarity}

A \emph{distance} is a function \(d: X \times X \to [0,+\infty[\). A \emph{distance metric} is a distance \(d\) that satisfies the following properties for all \(x, y, z \in X\):  
\begin{enumerate}
	\item \(d(x,x) = 0\) (identity)
    \item \(d(x, y) = 0 \implies x = y\) (separation), \label{separation}
    \item \(d(x, y) = d(y, x)\) (symmetry),
    \item \(d(x, z) \leq d(x, y) + d(y, z)\) (triangle inequality).
\end{enumerate}

A \emph{pseudo-distance metric} removes the separation property (\ref{separation}), allowing \(d(x, y) = 0\) for \(x \neq y\). This relaxation is particularly useful in supervised learning, where two distinct instances may share identical attribute values yet remain unequal.

A \textit{similarity relation} is a function \(R: X \times X \to [0,1]\), where larger values of \(R(x, y)\) indicate a greater resemblance between \(x\) and \(y\). For any given distance function \(d: X \times X \to [0,1]\), there exists a corresponding similarity measure \cite{deza2009encyclopedia}, defined as \(R(x, y) = 1 - d(x, y)\).

In this paper, we thus use the term \emph{distance} in a broad sense, without strictly adhering to the formal requirements of a (pseudo-) distance metric. Specifically, we impose only symmetry and identity, focusing instead on its primary role: quantifying the dissimilarity between instances.
\section{Choquet distances}
\label{sec: FRdistances}
A \emph{decision system} $\left( X, \mathcal{A}\cup \{d\}\right)$, consists of a finite non-empty set of instances \(X\), a non-empty family of conditional attributes $\mathcal{A}$ and a decision attribute \(d\notin \mathcal{A}\), where each attribute \(a\in \mathcal{A}\cup\{d\}\) is a function $a:\ X \rightarrow V_a$, with $V_a$ the set of values the attribute $a$ can take.
Throughout this section, we will assume $\left( X, \mathcal{A}\cup \{d\}\right)$ is a decision system.
\subsection{Choquet distances}
\label{sec: general Choquet distance}
For simplicity of notation and due to the effectiveness of the Manhattan distance in supervised learning \cite{lenz2023unified}, this paper focuses exclusively on the Manhattan-based variant of the Choquet distance. However, all the results in this paper can be easily adapted to a Minkowski \(p\)-distance version of the Choquet distance \cite{theerens2024fuzzy}.
\begin{definition}[Choquet distance]
	Suppose \(\mu\) is a monotone measure on the set of conditional attributes \(\mathcal{A}\).
	We define the Choquet distance \(d^\mu\) with respect to the monotone measure \(\mu\) as follows for \(x,y\in X\):
	\begin{equation}
		\label{eq: choquet distance}
		d^\mu(x,y) := \int |a(x)-a(y)| \diff \mu(a).
	\end{equation}
\end{definition}
\begin{remark}
	In this definition, we assume all conditional attributes to be numerical and the attribute-wise distance to be the Manhattan distance. However, this definition can easily be adjusted to accommodate other types of attributes and/or distances as well:
	\begin{equation*}
		d^{\mu}(x,y) := \int d_a(x,y) \diff \mu(a),
	\end{equation*}
	where \(d_a\) is a distance chosen for attribute \(a\).
\end{remark}
\begin{remark}
	Note that when we use an additive measure \(\nu\), Eq.\ \eqref{eq: choquet distance} turns into the weighted Manhattan distance (cf.\ Proposition \ref{prop: choqAdditive}):
\[d^\nu(x,y) = \sum_{a\in \mathcal{A}} \nu(\{a\})|a(x)-a(y)|.\] 
\end{remark}

\begin{example}
	\label{ex: 1}
	Consider the decision system illustrated in Table \ref{table:1}, presenting data on four patients, where the conditional attributes are fever, fatigue, and cough (each ranging from 0 to 1) and the decision attribute is the presence of a common cold.
	\begin{table}[H]
		\centering
		\def\arraystretch{1.1}%
		\setlength\tabcolsep{1.8 mm}
		\begin{tabular}{l |c |c| c ||c}
			        & \(a_1\) (fever) & \(a_2\) (fatigue) & \(a_3\) (cough) & \(d\) (common cold) \\
			\hline
			\(x_1\) & 0               & 0.9               & 0.9             & 1           \\
			\(x_2\) & 0.9             & 0.95              & 0.95            & 1           \\
			\(x_3\) & 0               & 1                 & 0               & 0           \\
			\(x_4\) & 0.9             & 0                 & 0               & 0
		\end{tabular}
		\caption{Decision system}
		\label{table:1}
	\end{table}
	\begin{flushleft}
		Suppose we want to assign the following weights to each attribute:  
	\end{flushleft}

\begin{equation}
    \label{eq: weights}
    w_{a_1} = 0.2, \quad w_{a_2} = 0.4, \quad w_{a_3} = 0.4.
\end{equation}  
One way to incorporate these weights is by using the weighted Manhattan distance, which corresponds to the Choquet distance with respect to the additive measure defined as \(\mu_w(A) = \sum_{a \in A} w_a\). Another option is to use the following measure:  
\begin{equation*}
    \begin{gathered}
        \mu(\{a_1\}) = \frac{w_{a_1}}{2} = 0.1, \quad \mu(\{a_2\}) = \frac{w_{a_2}}{2} = 0.2, \quad \mu(\{a_3\}) = \frac{w_{a_3}}{2} = 0.2, \\ 
        \mu(\{a_1, a_2\}) = 0.2, \quad \mu(\{a_1, a_3\}) = 0.2, \quad \mu(\{a_2, a_3\}) = 0.5, \\ 
        \mu(\mathcal{A}) = 1, \quad \mu(\emptyset) = 0.
    \end{gathered}
\end{equation*}  
By calculating the Shapley value\footnote{The Shapley value \cite{shapley1951notes} of an attribute \(a \in \mathcal{A}\) with respect to \(\mu\) is defined as:  
\[
\phi_\mu(a) = \sum_{A \subseteq \mathcal{A} \setminus \{a\}} \frac{|A|!(|\mathcal{A}| - |A| - 1)!}{|\mathcal{A}|!} \big[\mu(A \cup \{a\}) - \mu(A)\big],
\]  
which quantifies \(a\)'s contribution to the measure \(\mu\) over all subsets \(A \subseteq \mathcal{A}\).} for each attribute, we find that this measure assigns each attribute the same weight as specified in Equation \eqref{eq: weights}:
\[ \phi_\mu(a_1) = 0.2, \quad \phi_\mu(a_2) = 0.4, \quad \phi_\mu(a_3) = 0.4.\]

Table \ref{table: 2} compares the distances calculated using the Choquet distance with respect to the measure \(\mu\), the counting measure (i.e., the Manhattan distance; see Example \ref{exmp: counting}), and the additive measure \(\mu_w\) (i.e., the weighted Manhattan distance based on the weights from Equation \eqref{eq: weights}). As an example, let us calculate the Choquet distance between \(x_3\) and \(x_4\) (cf.\ Definition \ref{defn: ChoquetIntegral}). First, we calculate the attribute-wise distances:
\[
|a_1(x_3) - a_1(x_4)| = 0.9, \quad |a_2(x_3) - a_2(x_4)| = 1, \quad |a_3(x_3) - a_3(x_4)| = 0.
\]
Next, we sort the attributes \(\mathcal{A}\) into \((a^\ast_1, a^\ast_2, a^\ast_3)\) such that
\[
|a^\ast_1(x_3) - a^\ast_1(x_4)| \leq |a^\ast_2(x_3) - a^\ast_2(x_4)| \leq |a^\ast_3(x_3) - a^\ast_3(x_4)|.
\]
Hence, \(a^\ast_1 = a_3\), \(a^\ast_2 = a_1\), and \(a^\ast_3 = a_2\). Now we calculate the Choquet distance as:
\begin{align*}
    d^\mu(x_3, x_4) &= \int |a(x_3) - a(x_4)| \diff\mu(a) \\
    &= \sum_{i=1}^3 |a^\ast_i(x_3) - a^\ast_i(x_4)| \cdot \left[\mu(\{a^\ast_i, \dots, a^\ast_3\}) - \mu(\{a^\ast_{i+1}, \dots, a^\ast_3\})\right] \\
    &= 0.9 \cdot (\mu(\{a^\ast_2, a^\ast_3\}) - \mu(\{a^\ast_3\})) + 1 \cdot (\mu(\{a^\ast_3\}) - 0) \\
    &= 0.9 \cdot (0.2 - 0.2) + 1 \cdot 0.2 = 0.2.
\end{align*}

	\begin{table}[!htb]
		\def\arraystretch{1.1}%
		\setlength\tabcolsep{0.9 mm}
		\begin{subtable}{.33\linewidth}
			\centering
			\begin{tabular}{l |c |c| c| c}
				\(d^\mu\)& \(x_1\) & \(x_2\) & \(x_3\) & \(x_4\) \\
				\hline
				\(x_1\) & 0.0 & 0.135 & 0.21 & 0.9 \\
				\(x_2\) & 0.135 & 0.0 & 0.23 & 0.475 \\
				\(x_3\) & 0.21 & 0.23 & 0.0 & 0.2 \\
				\(x_4\) & 0.9 & 0.475 & 0.2 & 0.0 \\
			\end{tabular}
			\caption{Non-additive measure \(\mu\)}
		\end{subtable}%
		\begin{subtable}{.33\linewidth}
			\centering
			\begin{tabular}{l |c |c| c | c }
				\(d^\#\) & \(x_1\) & \(x_2\) & \(x_3\) & \(x_4\) \\
				\hline
				\(x_1\)    & 0.0     & 0.33    & 0.33    & 0.9     \\
				\(x_2\)    & 0.33    & 0.0     & 0.63    & 0.63    \\
				\(x_3\)    & 0.33    & 0.63    & 0.0     & 0.63    \\
				\(x_4\)    & 0.9     & 0.66    & 0.63    & 0.0
			\end{tabular}
			\caption{Counting measure \(\mu_\#\)}
		\end{subtable}%
		\begin{subtable}{.33\linewidth}
			\centering
			\begin{tabular}{l |c |c| c | c }
				\(d^w\) & \(x_1\) & \(x_2\) & \(x_3\) & \(x_4\) \\
				\hline
				\(x_1\)   & 0.0     & 0.22    & 0.4     & 0.9     \\
				\(x_2\)   & 0.22    & 0.0     & 0.58    & 0.76    \\
				\(x_3\)   & 0.4     & 0.58    & 0.0     & 0.58    \\
				\(x_4\)   & 0.9     & 0.76    & 0.58    & 0.0
			\end{tabular}
			\caption{Additive measure \(w\)}
		\end{subtable}
		\caption{Choquet distances with respect to several measures}
		\label{table: 2}
	\end{table}
	According to both \(d^\#\) and \(d^w\), \(x_1\) is identified as the nearest neighbor of \(x_3\). However, since \(x_1\) and \(x_3\) belong to different classes, this is an undesirable result. In contrast, the Choquet distance \(d^\mu\) correctly identifies \(x_4\) as the nearest neighbor of \(x_3\). This demonstrates that, although both the weighted distance \(d^w\) and the Choquet distance \(d^\mu\) assign the same weights to individual attributes, they produce different distance outcomes.
\end{example}

The procedure that details the computation of these Choquet distances (using Proposition \ref{prop: ChoquetIntegral2}) is presented in Algorithm \ref{alg:one}.

\begin{algorithm}[h]
	\SetNoFillComment
	\caption{Calculation of the Choquet distance}\label{alg:one}
	\KwData{Elements \(x,y \in \mathbb{R}^m\), a measure \(\mu:\pwset(\{1,2,\dots,m\})\to [0,1]\)}
	\KwResult{Choquet distance between \(x\) and \(y\) with respect to \(\mu\)}
	\tcc{Create list with the attribute-wise distances}
	distances $ \gets $[]\;
	\For{\(i \in \{1,2,\dots,m\}\)}{
		distances.append(\(\abs*{x_i-y_i}\))
	}
	\tcc{Sort the attribute-wise distances}
	sorted\_distances, sorted\_indices $\gets$ sortAscending(distances) \;
	$d^{\mu}(x,y) \gets $ sorted\_distances[$1$]\;
	\For{\(i \in \{2,3,\dots,m\}\)}{
	\tcc{Evaluate and store measure for later}
	current\_measure $\gets$ \(\mu\).evaluate(sorted\_indices\([i,\dots,m]\))\;
	current\_diff$\gets$ sorted\_distances[$i$] $-$ sorted\_distances[$i-1$]\;

	$d^{\mu}(x,y) \gets d^{\mu}(x,y) $ $ + $ current\_measure \(*\) current\_diff\;
	}
	\Return{$d^{\mu}(x,y)$}
\end{algorithm}
It is important to note that the measure \(\mu\) only needs to be evaluated for \(m\) subsets. By calculating and storing these values in Step 8, they can be reused for future distance calculations. This approach allows for an ``online'' computation of the measure, making the process more feasible. Without this optimization, i.e.\ precomputing the measure, the time complexity would be exponential in the number of attributes (\(O(2^m)\)).

The time complexity of the algorithm can be broken down as follows:
\begin{itemize}
	\item \textbf{Step 2:} Calculating the attribute-wise distances takes \(O(m)\)
	\item \textbf{Step 5:} Sorting takes \(O(m\log(m))\)
	\item \textbf{Step 7:} Iterating over \(m\) indices, with each iteration \(i\) requiring \(T_\mu(i)\), where \(T_\mu(i)\) denotes the time complexity of evaluating a subset of cardinality \(i\) in \(\mu\). The overall complexity for this step is \(O\left(\sum_{i=1}^m T_\mu(i)\right)\).

\end{itemize}

Therefore, the overall time complexity of calculating the Choquet distance is
\begin{equation}
	\label{eq: timeComplexityChoquetDistance}
	O\left(m\log(m) + \sum_{i=1}^m T_\mu(i)\right).
\end{equation}

The challenge now is to select an appropriate measure \(\mu\) for use in Eq.\ \eqref{eq: choquet distance} to define a concrete distance. In the absence of expert knowledge such as the medical information used in Example \ref{ex: 1}, one solution is to use dependency measures. Suppose \(\dep : \mathcal{P}(\mathcal{A} \cup \{d\}) \times \mathcal{P}(\mathcal{A} \cup \{d\}) \to [0,1]\) is a dependency measure, i.e., a function where \(\dep(A, B)\) quantifies the dependency of a set of attributes \(A\) on a set of attributes \(B\). We define the following measure on the conditional attribute space \(\mathcal{A}\):
\[\mu_{\dep}(B) := \dep(\{d\}, B), \;\; \forall B \in \mathcal{A}.\]
These measures quantify the importance of attribute subsets in predicting the decision attribute, making them suitable for supervised learning.

\section{Fuzzy-rough attribute importance measures}
\label{sec: attribute importance measures}
In this section, we introduce attribute importance measures based on dependency measures from fuzzy rough set theory. These measures can be used in Choquet distances for supervised learning. Additionally, we will analyze their time complexity.

Consider a family of similarity relations \(\{R_B: B\subseteq \mathcal{A}\}\), and a similarity relation \(R_d\) (\(R_B,R_d: X^2 \to [0,1]\)) for the decision attribute. The authors in \cite{cornelis2010attribute} define the $B$-positive region $POS_B$ as the fuzzy set\footnote{A fuzzy set \cite{zadeh1996fuzzy} in \(X\) is a function \(X\to [0,1]\).} in $X$ defined as, for $y \in X$,
$$POS_{R_B}(y) = \max_{x\in X}\min\limits_{z\in X} \imp(R_B(y,z),R_d(x,z)),$$
where \(\imp\) is an implicator\footnote{An \emph{implicator} is a binary operator $\imp: \left[0,1\right]^2\rightarrow \left[0,1\right]$ that is non-increasing in the first argument, non-decreasing in the second argument and for which $\imp(0,0)=\imp(0,1)=\imp(1,1)=1$ and \(\imp(1,0)=0\) holds.}.
The value \(POS_{R_B}(y)\) can be interpreted as the degree to which similarity with respect to the conditional attributes \(B\) relates to similarity with respect to the decision attribute.
Consequently, the predictive ability of a  subset \(B\) to predict the decision attribute \(d\), also called the degree of dependency of \(d\) on \(B\), is defined as:
$$\gamma_R(B) =\sum\limits_{y \in X}POS_{R_B}(y).$$

A second measure considers the worst case scenario, i.e., to what extent there exists an element totally outside of the positive region:
$$\delta_R(B) = \min_{y \in X}POS_{R_B}(y).$$
In the case of classification, we can simplify the positive region as follows (cf.\ \cite{theerens2024fuzzy}):
\begin{align*}
	POS_{d_B}(y)=\min\limits_{z \notin R_dy}d_B(z,y),
\end{align*}
where \(\{d_B: B\subseteq \mathcal{A}\}\) is a family of distances. This leads us to generalize our \(\gamma\) and \(\delta\) measure as follows:
\begin{equation}
	\label{eq: gammadelta}
	\gamma_d(B) =  \sum\limits_{y \in X}\min\limits_{z \notin R_dy}d_B(z,y)\;\text{ and }\;
	\delta_d(B) =\min\limits_{y \in X}\min\limits_{z \notin R_dy}d_B(z,y).
\end{equation}
The interpretation of Eq.\ \eqref{eq: gammadelta} is that the dependency of the decision \(d\) on a conditional attribute subset \(B\) can be interpreted as the average (or minimum in the case of \(\delta\)) of the distances between each instance and its closest neighbor from a different class.

This more general definition makes it easier to construct measures for classification:
\begin{example}
	\label{ex: gamma measure}
	Using the Chebyshev distance \(d_B(x,y) = \max\limits_{a\in B}|a(z)-a(y)|\), we get
	\begin{equation}
		\label{eq: gamma measure}
		\gamma_d(B) = \sum\limits_{y \in X}\min\limits_{z \notin R_dy}\left(\max\limits_{a\in B}|a(z)-a(y)|\right).
	\end{equation}
	The Chebyshev distance can, of course, be replaced with any other distance metric. By applying Equation \eqref{eq: gamma measure} to the decision system in Example \ref{ex: 1}, we obtain:
	\begin{equation}
		\label{eq: gammaconcrete}
		\begin{gathered}
			\gamma_d(\{a_1\})=0.0, \gamma_d(\{a_2\})=1.1 \text{ and } \gamma_d(\{a_3\})=3.65, \\ \gamma_d(\{a_1,a_2\})=2.0, \gamma_d(\{a_1,a_3\})=3.65, \gamma_d(\{a_2,a_3\})=3.65,\\
			\gamma_d(\mathcal{A})=3.65, \gamma_d(\emptyset)=0.
		\end{gathered}
	\end{equation}
	Note that this measure is not normalized (i.e., \(\mu(\mathcal{A}) \neq 1\)); however, this does not affect its usefulness, as we are only concerned with relative distances. 

As an example, let us calculate \(\gamma_d(\{a_1\})\): 

	\begin{align*}
		\gamma_d(\{a_1\}) &= \sum\limits_{y \in X}\min\limits_{z \notin R_dy}\left(\max\limits_{a\in \{a_1\}}|a(z)-a(y)|\right) = \sum\limits_{y \in X}\min\limits_{z \notin R_dy}\left(|a_1(z)-a_1(y)|\right) \\
		&= 2*|0 - 0| + 2*|0.9-0.9| = 0.
	\end{align*}
	The Choquet distances calculated using the \(\gamma\) measure from Eq. \eqref{eq: gammaconcrete} are displayed in Table \ref{table:5}.
	\begin{table}[H]
		\centering
		\def\arraystretch{1.1}%
		\setlength\tabcolsep{0.9 mm}
		\begin{tabular}{l |c |c| c | c }
			\(d^{\gamma_d}\) & \(x_1\) & \(x_2\) & \(x_3\) & \(x_4\) \\
			\hline
			\(x_1\) & 0.00 & 0.18 & 3.28 & 3.29 \\
        \(x_2\) & 0.18 & 0.00 & 3.47 & 3.47 \\
        \(x_3\) & 3.28 & 3.47 & 0.00 & 1.91 \\
        \(x_4\) & 3.29 & 3.47 & 1.91 & 0.00 \\
		\end{tabular}
		\caption{Fuzzy rough Choquet distance using \(\gamma_d\)}
		\label{table:5}
	\end{table}
	Compared with the distance used in Example \ref{ex: 1}, this distance brings instance of the same class closer together while pushing instances of different classes farther apart.
\end{example}

Next, we determine the time complexity of computing Choquet distances using these measures. Based on the discussion in Algorithm \ref{alg:one}, the time complexity is expressed by Eq.\ \eqref{eq: timeComplexityChoquetDistance}. The remaining task is to calculate \(T_\mu(i)\), which represents the time complexity of evaluating a subset of cardinality \(i\) in \(\mu\).
First, consider the case where the decision attribute is continuous, i.e., we are working in a regression setting. In this case, we want to calculate:
\begin{align}
	\label{eq: gamma Supervised}
	\gamma_R(B) = \sum\limits_{y \in X}\max\limits_{x\in X}\min\limits_{z\in X} \imp(R_B(y,z),R_d(x,z)).
\end{align}
Therefore, considering the three nested loops of size \(n = |X|\) and assuming the time complexity of calculating \(R_B\) is \(O(|B|)\) (which is the case for all relations used in this paper), we obtain:
\[T_\gamma(i)= O(n^3 * i).\]
 The total time complexity for calculating the Choquet distance between two points with respect to \(\gamma\) (cf.\ Eq.\ \eqref{eq: timeComplexityChoquetDistance}) is:
\begin{align*}
	O\left(m\log(m) + \sum_{i=1}^m T_\gamma(i)\right) & = O\left(m\log(m) + \sum_{i=1}^m n^3 * i\right)=O\left(m\log(m) + n^3 * \frac{m*(m+1)}{2}\right) \\
	                                                  & = O(n^3*m^2).
\end{align*}
Analogous reasoning applies to the time complexity for \(\delta\).\\
In the case the decision attribute is categorical, i.e.\ we are in a classification setting, we want to calculate:
\[\gamma_d(B) =  \sum\limits_{y \in X}\min\limits_{z \notin R_dy}d_B(z,y).\]
Therefore, we only have two nested loops in this case, resulting in the following time complexity for calculating the Choquet distance between two points:
\begin{align*}
	O\left(m\log(m) + \sum_{i=1}^m T_\gamma(i)\right) & = O\left(m\log(m) + \sum_{i=1}^m n*O(n) * i\right)= O(n^2*m^2).
\end{align*}

\section{Choquet similarities and their duality to Choquet distances}
\label{sec: choquet similarities}
In this section, for simplicity, we assume that all attributes \(a\) have been normalized, meaning \(a: X \to [0,1]\) for every \(a \in \mathcal{A}\).
Instead of defining distances, we could have defined similarities in exactly the same way.
\begin{definition}[Choquet similarity]
	Suppose \(\mu\) is a monotone measure on the set of conditional attributes \(\mathcal{A}\).
	We define the Choquet similarity \(R^\mu\) with respect to the monotone measure \(\mu\) as follows:
	\begin{equation}
		\label{eq: choquet similarity}
		R^{\mu}(x,y) := \int 1- |a(x)-a(y)|\diff \mu(a).
	\end{equation}
\end{definition}
\begin{remark}
	\label{remark: choqSimilar}
    For general attributes, the Choquet similarity can be defined as:
    \[
    R^{\mu}(x,y) := \int 1 - \frac{|a(x) - a(y)|}{D} \, \mathrm{d}\mu(a),
    \]
    where \(D = \max_{a \in \mathcal{A}} D_a\), with \(D_a\) being an upper bound for the attribute-wise distance of attribute \(a\).
\end{remark}
Let us take a closer look at the definition of Choquet similarity:
\begin{align*}
	R^\mu(x,y) & =\int1- \abs{a(x)-a(y)} \diff \mu(a)=\sum_{i=1}^n \left(1-\abs{a^\ast_i(x)-a^\ast_i(y)}\right)\cdot\left[\mu(A^\ast_i)-\mu(A^\ast_{i+1})\right] \\
	           & =\sum_{i=1}^n \left(1-\abs{a^\ast_i(x)-a^\ast_i(y)}\right)\cdot\left[\mu(A^\ast_i)-\mu(A^\ast_{i}\backslash\left\{a^\ast_i\right\})\right],
\end{align*}
where \((a^\ast_1,a^\ast_2,\dots,a^\ast_n)\) is a permutation of \(\mathcal{A}=(a_1,a_2,\dots,a_n)\) such that
\begin{equation*}
	1-\abs{a^\ast_1(x)-a^\ast_1(y)}\leq 1-\abs{a^\ast_2(x)-a^\ast_2(y)} \leq\cdots\leq 1-\abs{a^\ast_n(x)-a^\ast_n(y)},
\end{equation*}
\(A^\ast_i:=\{a^\ast_i,\dots,a^\ast_n\}\) and \(\mu(A^\ast_{n+1}):=0\). Thus, when calculating the Choquet similarity between two instances \(x\) and \(y\), we first order the conditional attributes such that the similarity with respect to the \(i\)th attribute is the \(i\)th smallest. The similarity with respect to this \(i\)th attribute is then weighted by \(\mu(A^\ast_i)-\mu(A^\ast_{i}\backslash\left\{a^\ast_i\right\})\). Conversely, when calculating the Choquet distance, we order the conditional attributes such that the distance is the \(i\)th smallest—in other words, the inverse ordering used for calculating the Choquet similarity. This causes the same measure to weight our attributes differently, resulting in the lack of classic symmetry between distance and similarity, which is present in the weighted distance/similarity case.

The asymmetry between the Choquet distance and Choquet similarity arises from the fact that, when considering the dual measure \(\overline{\mu}\):  
\[
\overline{\mu}(B) := \mu(\mathcal{A}) - \mu(\mathcal{A} \setminus B), \quad \forall B \subseteq \mathcal{A},
\]  
where \(\mu\) describes the importance of a subset of conditional attributes, we still obtain a valid attribute importance measure. Indeed, if \(\mu(B)\) quantifies how important (or beneficial) a subset \(B\) is for predicting the decision attribute, then \(\overline{\mu}(B)\) represents how much worse the prediction becomes without \(B\). This, in turn, also reflects the importance of \(B\).

The following proposition illustrates the precise nature of the asymmetry between Choquet distance and Choquet similarity.
\begin{proposition}
	\label{prop: dualityChoquetdistance}
	Suppose \(\mu\) is a monotone measure, then we have the following:
	\[R^\mu (x,y)= \mu(\mathcal{A})- d^{\overline{\mu}}(x,y),\]
	where \(\overline{\mu}(B) := \mu(\mathcal{A}) - \mu(\mathcal{A}\bslash B)\) is the dual measure of \(\mu\).
\end{proposition}
\begin{remark}
    For general attributes, this proposition takes the form (using the notation from Remark \ref{remark: choqSimilar}):  
    \[
    R^\mu(x, y) = \mu(\mathcal{A}) - \frac{d^{\overline{\mu}}(x, y)}{D}.
    \]
\end{remark}
\begin{proof}
	Follows directly from Proposition \ref{choqintegralDual}:
	\begin{align*}
		R^{\mu}(x,y) &= \int 1- |a(x)-a(y)|\diff \mu(a)= \mu(\mathcal{A})+ \int- |a(x)-a(y)|\diff \mu(a) \\
		&=  \mu(\mathcal{A})- \int |a(x)-a(y)|\diff \overline{\mu}(a) = \mu(\mathcal{A})- d^{\overline{\mu}}(x,y).
	\end{align*}
\end{proof}

This explains why weighted distances exhibit the classical symmetry between similarity and distance. Indeed, consider an additive measure \(\mu(B) = \sum_{a \in B} w_a\). Then, we have  
\[
\overline{\mu}(B) = \mu(\mathcal{A}) - \sum_{a\in \mathcal{A} \setminus B} w_a = \sum_{a \in \mathcal{A}} w_a - \sum_{a \in \mathcal{A}\setminus B} w_a = \sum_{a \in B} w_a = \mu(B),
\]
which shows that additive measures are self-dual (i.e., \(\overline{\mu} = \mu\)). Note, however, that the converse is not true; not every self-dual measure is additive. As Choquet distances with respect to additive measures are equivalent to weighted distances, the previous proposition guarantees the classical symmetry between weighted similarity and weighted distance.

\begin{cor}
	\label{cor: self-dual}
	Suppose \(\mu\) is a self-dual measure (i.e.\ \(\overline{\mu}= \mu\)), then we have
	\[R^\mu (x,y)= \mu(\mathcal{A}) - d^{\mu}(x,y).\]
\end{cor}

To regain the classical symmetry between distances and similarities we can symmetrize our measure by defining \(\mu^s = (\mu +\overline{\mu})/2\), giving us the following symmetric Choquet distance and Choquet similarity:
\begin{align}
	d^{\mu^s}(x,y):=&\int \abs{a(x)-a(y)} (\text{d}(\mu + \overline{\mu})/2)\label{eq: symmetric Choquet}\\ 
	=& \frac{1}{2}\left(\int \abs{a(x)-a(y)} \diff \mu + \int \abs{a(x)-a(y)} \diff \overline{\mu}\right),\nonumber\\
	R^{\mu^s}(x,y):=&\int 1-\abs{a(x)-a(y)} (\text{d}(\mu + \overline{\mu})/2)   \nonumber               \\
	=& \frac{1}{2}\left(\int 1-\abs{a(x)-a(y)} \diff \mu + \int 1-\abs{a(x)-a(y)} \diff \overline{\mu}\right),\nonumber
\end{align}
where the second equality in every definition can easily be seen from Proposition \ref{prop: ChoquetIntegral2}.
Using the symmetric Choquet distance and Choquet similarity we do indeed have the normal symmetry between distance and similarity:
\begin{proposition}
	Suppose \(\mu\) is a monotone measure on \(\mathcal{A}\), then we have the following:
	\[R^{\mu^s}(x,y)= \mu(\mathcal{A}) - d^{\mu^s}(x,y)\]
\end{proposition}
\begin{proof}
	Follows from Corollary \ref{cor: self-dual} and the fact that \(\mu^s\) is self-dual:
	\begin{align*}
		\overline{\mu^s}(B) & =\overline{\left(\frac{\mu+\overline{\mu}}{2}\right)}(B)= 1- \left(\frac{\mu+\overline{\mu}}{2}\right)(\mathcal{A}\bslash B)= 1- \frac{\mu(\mathcal{A}\bslash B)+ \overline{\mu}(\mathcal{A}\bslash B)}{2} \\
		                    & = \frac{2-\left(\mu(\mathcal{A}\bslash B)+ 1- \mu(B)\right)}{2}=\frac{1+\mu(B)-\mu(\mathcal{A}\bslash B)}{2} = \mu^s(B).
	\end{align*}
\end{proof}

To unify these different Choquet distances we introduce the \(\alpha\)-Choquet distance (\(\alpha\in[0,1]\)):
\begin{equation}
	\prescript{\alpha}{\mu}{d}(x, y) := \int |a(x) - a(y)| \diff \mu_{\alpha}(a), \quad \mu_{\alpha} = (1 - \alpha) \mu + \alpha \overline{\mu}. \label{eq: alhpaChoq}
\end{equation}
	We have the following special cases:
	\begin{itemize}
		\item When \(\alpha = 0\), \(\prescript{0}{\mu}{d}\) reduces to the standard Choquet distance \(d^\mu\).
		\item When \(\alpha = \frac{1}{2}\), \(\prescript{0.5}{\mu}{d}\) reduces to the symmetric Choquet distance \(d^{\mu^s}\).
		\item When \(\alpha = 1\), \(\prescript{1}{\mu}{d}\) becomes the distance corresponding with the Choquet similarity, i.e., \(\prescript{1}{\mu}{d} = \mu(\mathcal{A}) - R^\mu= d^{\overline{\mu}}\) (cf.\ Proposition \ref{prop: dualityChoquetdistance}).
	\end{itemize}
	\begin{example}
	Recalling Example \ref{ex: gamma measure}, we calculate \(\prescript{1}{\mu}{d}\) and \(\prescript{0.5}{\mu}{d}\). First, we calculate \(\overline{\gamma_d}\) and \(\gamma_d^s\):
	\begin{equation*}
		\begin{gathered}
			\overline{\gamma_d}(\{a_1\})=0.0, \overline{\gamma_d}(\{a_2\})=0.0 \text{ and } \overline{\gamma_d}(\{a_3\})=1.65, \\ \overline{\gamma_d}(\{a_1,a_2\})=0.0, \overline{\gamma_d}(\{a_1,a_3\})=2.55, \overline{\gamma_d}(\{a_2,a_3\})=3.65,\\
			\overline{\gamma_d}(\mathcal{A})=3.65, \overline{\gamma_d}(\emptyset)=0,
		\end{gathered}
	\end{equation*}
	\begin{equation*}
		\begin{gathered}
			\gamma_d^s(\{a_1\})=0.0, \gamma_d^s(\{a_2\})=0.55 \text{ and } \gamma_d^s(\{a_3\})=2.65, \\ \gamma_d^s(\{a_1,a_2\})=1.0, \gamma_d^s(\{a_1,a_3\})=3.1, \gamma_d^s(\{a_2,a_3\})=3.65,\\
			\gamma_d^s(\mathcal{A})=3.65, \gamma_d^s(\emptyset)=0,
		\end{gathered}
	\end{equation*}
	which gives us the \(\alpha\)-Choquet distances in Table \ref{table: alphaChoquetdistances}.

	\begin{table}[!htb]
		\def\arraystretch{1.1}%
		\setlength\tabcolsep{0.9 mm}
		\begin{subtable}{.33\linewidth}
			\centering
			\begin{tabular}{l |c |c| c| c}
				\(\prescript{0}{\mu}{d}\)& \(x_1\) & \(x_2\) & \(x_3\) & \(x_4\) \\
				\hline
				\(x_1\) & 0.00 & 0.18 & 3.28 & 3.29 \\
				\(x_2\) & 0.18 & 0.00 & 3.47 & 3.47 \\
				\(x_3\) & 3.28 & 3.47 & 0.00 & 1.91 \\
				\(x_4\) & 3.29 & 3.47 & 1.91 & 0.00 \\
			\end{tabular}
			\caption{Fuzzy rough measure \(\gamma_d\)}
		\end{subtable}%
		\begin{subtable}{.33\linewidth}
			\centering
			\begin{tabular}{l |c |c| c | c }
				\(\prescript{0.5}{\mu}{d}\) & \(x_1\) & \(x_2\) & \(x_3\) & \(x_4\) \\
				\hline
		\(x_1\) & 0.00 & 0.18 & 2.48 & 3.29 \\
        \(x_2\) & 0.18 & 0.00 & 2.95 & 3.47 \\
        \(x_3\) & 2.48 & 2.95 & 0.00 & 0.96 \\
        \(x_4\) & 3.29 & 3.47 & 0.96 & 0.00 \\
			\end{tabular}
			\caption{Self-dual measure \(\gamma_d^s\)}
		\end{subtable}%
		\begin{subtable}{.33\linewidth}
			\centering
			\begin{tabular}{l |c |c| c | c }
				\(\prescript{1}{\mu}{d}\) & \(x_1\) & \(x_2\) & \(x_3\) & \(x_4\) \\
				\hline
			\(x_1\) & 0.00 & 0.18 & 1.69 & 3.29 \\
        \(x_2\) & 0.18 & 0.00 & 2.43 & 3.47 \\
        \(x_3\) & 1.69 & 2.43 & 0.00 & 0.00 \\
        \(x_4\) & 3.29 & 3.47 & 0.00 & 0.00 \\
			\end{tabular}
			\caption{Dual measure \(\overline{\gamma_d}\)}
		\end{subtable}
		\caption{\(\alpha\)-Choquet distances for \(\alpha = 0, 0.5, 1\).}
		\label{table: alphaChoquetdistances}
	\end{table}
	\end{example}
\section{Choquet distances seen as feature subset weighted distances}
\label{sec: Choquet distances seen as feature subset weighted distances}
The \(\alpha\)-Choquet distance \( \prescript{\alpha}{\mu}{d}(x,y) \) can be expressed in terms of the Möbius transform as (cf.\ Proposition \ref{mobiusChoquet} and Proposition \ref{mobiusChoquetDual}):
\[\prescript{\alpha}{\mu}{d}(x, y) = \sum_{B \subseteq \mathcal{A}} \mathcal{M}_\mu (B)\left((1 - \alpha) \cdot d^{-}_B(x, y) + \alpha \cdot d^{+}_B(x, y)\right),\]
where \(d^{-}_B(x,y) := \min_{a \in B}|a(x)-a(y)|\) and \(d^{+}_B(x, y) := \max_{a \in B} |a(x) - a(y)|\). This formulation shows that the Choquet distance can be interpreted as a subset weighted distance. 
Note that defining a subset-weighted distance as:
\[
\sum_{B \subseteq \mathcal{A}} \mathcal{M}_\mu(B) \frac{\sum_{a \in B} |a(x) - a(y)|}{|B|} = \sum_{B \subseteq \mathcal{A}} \mathcal{M}_\mu(B) d_B(x, y),
\]
where \( d_B(x, y) = \sum_{a \in B} \left( \frac{|a(x) - a(y)|}{|B|} \right) \), is not effective. Although this expression resembles a feature subset-weighted distance, it is actually a weighted distance, as will be shown in the following proposition. Moreover, the computation of the weights of this weighted distance requires evaluating an exponential number of subsets, making this method impractical.

\begin{proposition}
    For any set function \( \hat{\mu}: \mathcal{P}(\mathcal{A}) \to \mathbb{R} \), the following holds:
    \[
        \sum_{B \subseteq \mathcal{A}} \hat{\mu}(B) \frac{\sum_{a \in B} f(a)}{|B|} = \sum_{a \in \mathcal{A}} w_a f(a), \quad \forall f: \mathcal{A} \to \mathbb{R},
    \]
    where the weights \(w_a\) are given by
    \[
        w_a = \sum_{B \subseteq \mathcal{A}; \, a \in B} \frac{\hat{\mu}(B)}{|B|}, \quad \forall a \in \mathcal{A}.
    \]
\end{proposition}

\begin{proof}
    Define the functional \(\mathcal{L}\) as
    \[
        \mathcal{L}(f) = \sum_{B \subseteq \mathcal{A}} \hat{\mu}(B) \frac{\sum_{a \in B} f(a)}{|B|}, \quad \forall f: \mathcal{A} \to \mathbb{R}.
    \]
    Since \(\mathcal{L}\) is clearly linear, it can be expressed as a weighted sum. To determine the weight \(w_{a'}\), consider the function \(f(a)\) defined as \(f(a) = 1\) if \(a = a'\) and \(f(a) = 0\) otherwise. Substituting this into \(\mathcal{L}(f)\) gives the desired result.
\end{proof}
\section{Duplicate feature robustness of Choquet distances}
\label{sec: duplicatesChoquet}
In this section, we investigate the simplest form of interaction between attributes: duplicates. We demonstrate that Choquet distances possess the notable property of robustness against adding duplicate features.

Suppose \(a\) is a duplicate of the feature \(b\). Intuitively, an attribute importance measure \(\mu\) should treat these two attributes equivalently, ensuring that adding \(a\) or \(b\) to a set results in the same value of the measure. More formally, we require \(\mu\) to satisfy the following condition:
\begin{definition}
    \label{defn:duplicates}
    Let \(\mu\) be a measure on \(\mathcal{A}\). We say that \(a, b \in \mathcal{A}\) are \textbf{duplicates with respect to} \(\mu\) if 
    \[
        \mu(A \cup \{a\}) = \mu(A \cup \{b\}) \quad \forall A \subseteq \mathcal{A}.
    \]
\end{definition}

\begin{proposition}
	\label{prop: equivDuplicates}
    Let \(\mu\) be a measure on \(\mathcal{A}\) and \(a, b \in \mathcal{A}\). Then, the following are equivalent:
    \begin{enumerate}
        \item \(a\) and \(b\) are duplicates with respect to \(\mu\),
        \item \(\mu(A) = \mu(A \cup \{a\}) = \mu(A \cup \{b\}) \quad \forall A \subseteq \mathcal{A} \text{ such that } a \in A \lor b \in A,\)
        \item \(\mu(A) = \mu(A \setminus \{a\}) = \mu(A \setminus \{b\}) \quad \forall A \subseteq \mathcal{A} \text{ such that } a, b \in A.\)
        \item \(\mu(A \setminus \{a\}) = \mu(A \setminus \{b\}) \quad \forall A \subseteq \mathcal{A} \text{ such that } a, b \in A.\)
    \end{enumerate}
\end{proposition}

\begin{proof}
    \begin{itemize}
        \item \((1) \Rightarrow (2)\): If \(a \in A\) (or \(b \in A\)), then \(\mu(A \cup \{a\}) = \mu(A) = \mu(A \cup \{b\})\).
        \item \((2) \Rightarrow (3)\): If \(a, b \in A\), then \(a \in A \setminus \{b\}\). Hence, 
        \[
            \mu(A \setminus \{b\}) = \mu((A \setminus \{b\}) \cup \{b\}) = \mu(A),
        \]
        and similarly, \(\mu(A) = \mu(A \setminus \{a\})\).
		\item \((3) \Rightarrow (4)\): Trivial.
        \item \((4) \Rightarrow (1)\): If \(A \subseteq \mathcal{A}\), then \(a, b \in A \cup \{a, b\}\). Therefore,
        \[
            \mu(A \cup \{a\})= \mu((A \cup \{a, b\})\setminus \{b\}) =  \mu((A \cup \{a, b\})\setminus \{a\})  = \mu(A \cup \{b\}).
        \]
    \end{itemize}
\end{proof}
The following proposition demonstrates that the Choquet integral with respect to a measure for which \(a\) and \(b\) are two duplicate features can be simplified to a Choquet integral where one of these duplicate features are removed.
\begin{proposition}
	\label{prop: duplicateChoquet}
    Let \(\mu\) be a measure on \(\mathcal{A}\), and let \(a, b \in \mathcal{A}\) be duplicates with respect to \(\mu\). Suppose \(f: \mathcal{A} \to \mathbb{R}\) satisfies \(f(a) \leq f(b)\). Then, we have:
    \[
    \int f \, \mathrm{d}\mu = \int f|_{\mathcal{A} \setminus \{a\}} \, \mathrm{d} \mu|_{\mathcal{P}(\mathcal{A} \setminus \{a\})},
    \]
    where \(f|_{\mathcal{A} \setminus \{a\}}\) denotes the restriction of \(f\) to \(\mathcal{A} \setminus \{a\}\), and \(\mu|_{\mathcal{P}(\mathcal{A} \setminus \{a\})}\) denotes the restriction of \(\mu\) to subsets of \(\mathcal{A} \setminus \{a\}\).
\end{proposition}
\begin{proof}
    Define \(a^\ast_i\) and \(A^\ast_i\) as in Definition \ref{defn: ChoquetIntegral}. Now suppose \(a = a^\ast_j\) and \(b = a^\ast_k\). Since \(f(a) \leq f(b)\), we have, without loss of generality, \(j < k\). Using the definition of the Choquet integral (Definition \ref{defn: ChoquetIntegral}) and the fact that \(a\) and \(b\) are duplicates, we proceed as follows:
    \begin{align*}
        \int f \, \mathrm{d}\mu 
        & = \sum_{i=1}^n f(a^\ast_i) \cdot \left[\mu(A^\ast_i) - \mu(A^\ast_{i+1})\right] \\
        & = \sum_{i=1}^n f(a^\ast_i) \cdot \left[\mu(A^\ast_i) - \mu(A^\ast_i \setminus \{a^\ast_i\})\right] \\
        & = \left(\sum_{i \neq j} f(a^\ast_i) \cdot \left[\mu(A^\ast_i) - \mu(A^\ast_i \setminus \{a^\ast_i\})\right]\right) 
        + f(a^\ast_j) \cdot \left[\mu(A^\ast_j) - \mu(A^\ast_j \setminus \{a^\ast_j\})\right] \\
        & = \sum_{i < j} f(a^\ast_i) \cdot \left[\mu(A^\ast_i) - \mu(A^\ast_i \setminus \{a^\ast_i\})\right] 
        + \sum_{j < i} f(a^\ast_i) \cdot \left[\mu(A^\ast_i) - \mu(A^\ast_i \setminus \{a^\ast_i\})\right] \\
        & = \sum_{i < j} f(a^\ast_i) \cdot \left[\mu(A^\ast_i \setminus \{a\}) - \mu(A^\ast_i \setminus \{a^\ast_i, a\})\right] \\
        & \quad + \sum_{j < i} f(a^\ast_i) \cdot \left[\mu(A^\ast_i \setminus \{a\}) - \mu(A^\ast_i \setminus \{a^\ast_i, a\})\right] \\
        & = \int f|_{\mathcal{A} \setminus \{a\}} \, \mathrm{d} \mu|_{\mathcal{P}(\mathcal{A} \setminus \{a\})}.
    \end{align*}
    In the second-to-last step, we used the fact that for \(j < i\), \(a\) is not contained in \(A^\ast_i\), and for \(i < j\), both \(A^\ast_i\) and \(A^\ast_i \setminus \{a^\ast_i\}\) contain \(a\) and \(b\). By Proposition \ref{prop: duplicateChoquet}, this leads to the required equality.
\end{proof}
Applying this proposition to Choquet distances, we have that if \(a\) and \(b\) are duplicates with respect to \(\mu\), then the Choquet distance remains unchanged when one of the duplicate features is removed. An example of this is provided in Section \ref{sec: synth}.
Moreover, we note that duplication is invariant under duality, implying that the Choquet similarity also remains unchanged when duplicate features are removed:
\begin{proposition}
	\label{prop:dualduplicate}
	Let \(\mu\) be a measure on \(\mathcal{A}\). The elements \(a, b \in \mathcal{A}\) are duplicates w.r.t.\ \(\mu\) if and only if they are duplicates w.r.t.\ \(\overline{\mu}\).
\end{proposition}
\begin{proof}
  Suppose \(a, b \in \mathcal{A}\) are duplicates with respect to \(\mu\), and let \(A \subseteq \mathcal{A}\) such that \(a, b \in A\):  
\begin{align*}
  \overline{\mu}(A \setminus \{a\}) &= \mu(\mathcal{A}) - \mu(\mathcal{A} \setminus (A \setminus \{a\})) \\
  &= \mu(\mathcal{A}) - \mu((\mathcal{A} \setminus A) \cup \{a\}) \\
  &= \mu(\mathcal{A}) - \mu((\mathcal{A} \setminus A) \cup \{b\}) \\
  &= \overline{\mu}(A \setminus \{b\}),
\end{align*}
thus, by Proposition~\ref{prop: equivDuplicates}, \(a\) and \(b\) are also duplicates with respect to \(\overline{\mu}\).  
The converse follows from the idempotency of taking the dual, i.e., \(\overline{\overline{\mu}} = \mu\).
\end{proof}
\section{Experiments}
\label{sec: experiments}
In this section, we evaluate the proposed Choquet distances by comparing them to weighted distances and Mahalanobis distances. This evaluation is conducted on both a synthetic dataset, to assess robustness against duplicates and highly correlated features, and benchmark UCI datasets, to evaluate overall performance.  

\subsection{Evaluated distances}
Below, we describe each distance used in his section in detail.

\paragraph{Manhattan Distance (MAN)}
The Manhattan distance is defined as:
\begin{equation*}
    d_{\text{MAN}}(x, y) = \sum_{a \in \mathcal{A}} |a(x) - a(y)|.
\end{equation*}
It treats all features equally and is less sensitive to outliers than the Euclidean distance.

\paragraph{\(\chi^2\)-Weighted Manhattan Distance (CHI)}
The \(\chi^2\)-weighted Manhattan distance incorporates feature importance based on the \(\chi^2\) statistic:
\begin{equation*}
    d_{\text{CHI}}(x, y) = \sum_{a \in \mathcal{A}} w_a |a(x) - a(y)|,
\end{equation*}
where \(w_a\) corresponds to the \(\chi^2\) statistic for feature \(a\), quantifying its relevance to the decision attribute.

\paragraph{Mutual Information Weighted Manhattan Distance (MI)}
In this variant, feature weights are determined by mutual information between each feature and the decision attribute:
\begin{equation}
	\label{eq: MIdistance}
    d_{\text{MI}}(x, y) = \sum_{a \in \mathcal{A}} I(a; d) |a(x) - a(y)|,
\end{equation}
where \(I(a; d)\) denotes the mutual information between attribute \(a\) and decision attribute \(d\).

\paragraph{Mahalanobis Distance (MAH)}
Mahalanobis distance accounts for feature correlations and is defined as \cite{deza2009encyclopedia,mahalanobis2018generalized}:
\begin{equation*}
    d_{\text{MAH}}(x, y) = \sqrt{(\mathbf{a}(x) - \mathbf{a}(y))^T \mathbf{\Sigma}^{-1} (\mathbf{a}(x) - \mathbf{a}(y))},
\end{equation*}
where \(\mathbf{\Sigma}\) is the covariance matrix of the dataset and \(\mathbf{a}(x)\in\mathbb{R}^m\) is the attribute vector of an instance \(x\).

\paragraph{Mahalanobis Manhattan Distance (MAH\textsubscript{1})}
Since all evaluated distances, including the Choquet distances, are based on the Manhattan distance, we also consider a Manhattan variant of the Mahalanobis distance. This method first applies the \emph{whitening transformation} from the Mahalanobis distance, followed by the Manhattan distance:
\begin{equation}
	\label{eq: mahalanobis manhattan}
    d_{\text{MAH}_1}(x, y) = \sum_{i=1}^m \left| (\mathbf{\Sigma}^{-\frac{1}{2}} \mathbf{a}(x))_i - (\mathbf{\Sigma}^{-\frac{1}{2}} \mathbf{a}(y))_i \right|.
\end{equation}
This ensures the distance accounts for feature correlations while preserving the robustness of Manhattan distance.

\paragraph{Mutual Information Weighted Mahalanobis Manhattan Distance (MAMI)}
Given that many evaluated distances incorporate feature importance, we also consider a mutual information (MI)-weighted variant of the Mahalanobis distance. This method extends MAH\textsubscript{1} by first applying the whitening transformation on the training set and then computing mutual information weights in the transformed space. These weights are subsequently used to weight the Mahalanobis Manhattan distance.

\paragraph{Fuzzy Rough \(\gamma\)-Weighted Distance (WFR)}
This method assigns feature weights based on the fuzzy rough \(\gamma\)-measure:
\begin{equation*}
    d_{\text{WFR}}(x, y) = \sum_{a \in \mathcal{A}} \gamma_d(\{a\}) |a(x) - a(y)|,
\end{equation*}
where \(\gamma_d\) is defined in Equation \eqref{eq: gamma measure}.
\paragraph{\(\alpha\)-Choquet distance with \(\alpha \in \{0, 0.5, 1\}\) (CFR, CFR\textsubscript{.5}, CFR\textsubscript{1})}
The proposed \(\alpha\)-Choquet distance, defined in Equation~\eqref{eq: alhpaChoq} for \(\alpha \in \{0, 0.5, 1\}\), employing \(\gamma_d\) from Equation~\eqref{eq: gamma measure} as the underlying measure.
\subsection{Experiment: synthetic dataset}
\label{sec: synth}
In this subsection, we evaluate the performance of the Choquet distance and its robustness to duplicate features on synthetic datasets, comparing it to weighted distances and the Mahalanobis distance. Additionally, we examine its behavior in the presence of strongly correlated features.
\subsubsection{Construction and results}
To evaluate the performance of different distance measures when adding duplicates and highly correlated features, we perform K-Nearest Neighbors (KNN) classification with \(K = 5\). As we will see, the specific variation of the weighted, Mahalanobis, or Choquet distance is not crucial; only the type itself matters, as the results remain consistent across all variations. Therefore, to enhance clarity and improve the interpretability of the plots, we focus on the following four representative distances:  
\begin{itemize}  
    \item Manhattan distance (MAN),  
    \item Mutual information-weighted Manhattan distance (MI),  
    \item Mahalanobis Manhattan distance (MAH\textsubscript{1}),  
    \item FR-symmetric Choquet distance (CFR\textsubscript{.5}).  
\end{itemize}  

\begin{figure}[H]
    \centering
    \begin{subfigure}{\textwidth}
        \centering
        \includegraphics[width=\textwidth]{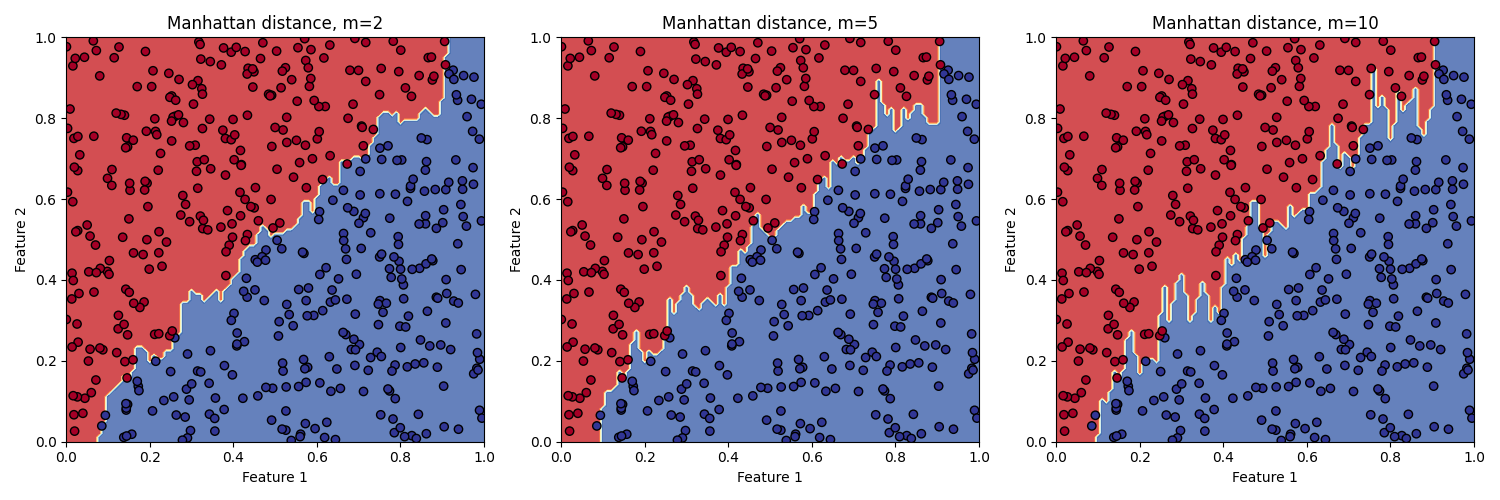} 
        \caption{Manhattan distance for \(m=2,5,10\)}
    \end{subfigure}
    
    \begin{subfigure}{\textwidth}
        \centering
        \includegraphics[width=\textwidth]{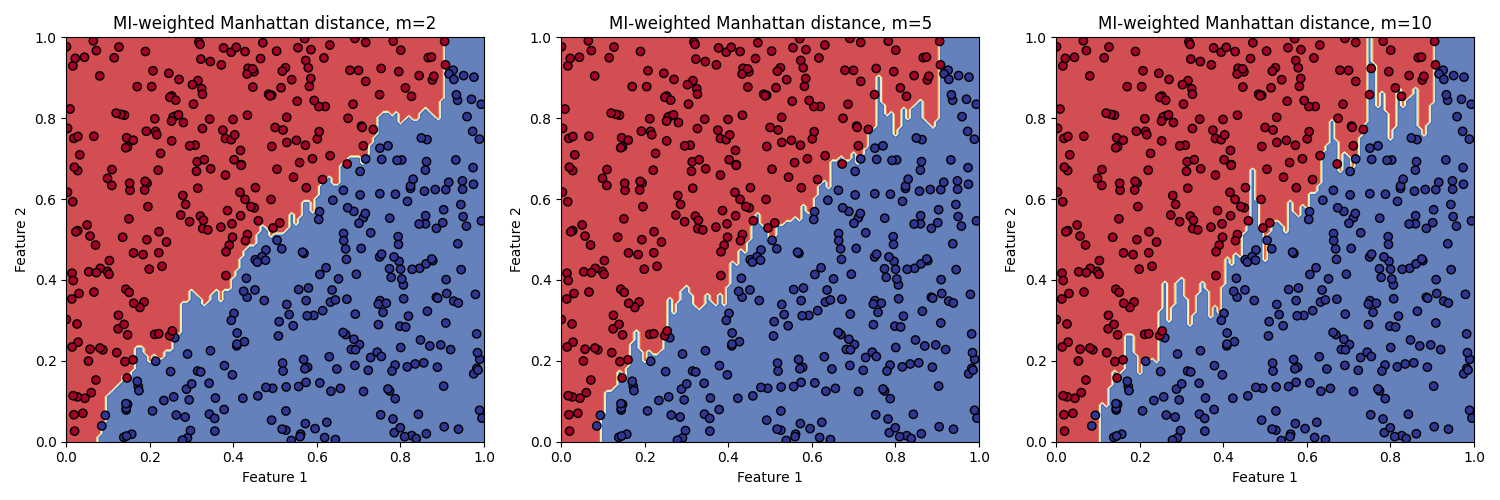}
        \caption{MI-weighted Manhattan distance for \(m=2,5,10\)}
    \end{subfigure}
    \caption{Classification boundaries of the first two distances in a 5-NN setting.}
    \label{fig:visualBoundary1}
\end{figure}

\begin{figure}[ht!]
	\begin{subfigure}{\textwidth}
        \centering
        \includegraphics[width=\textwidth]{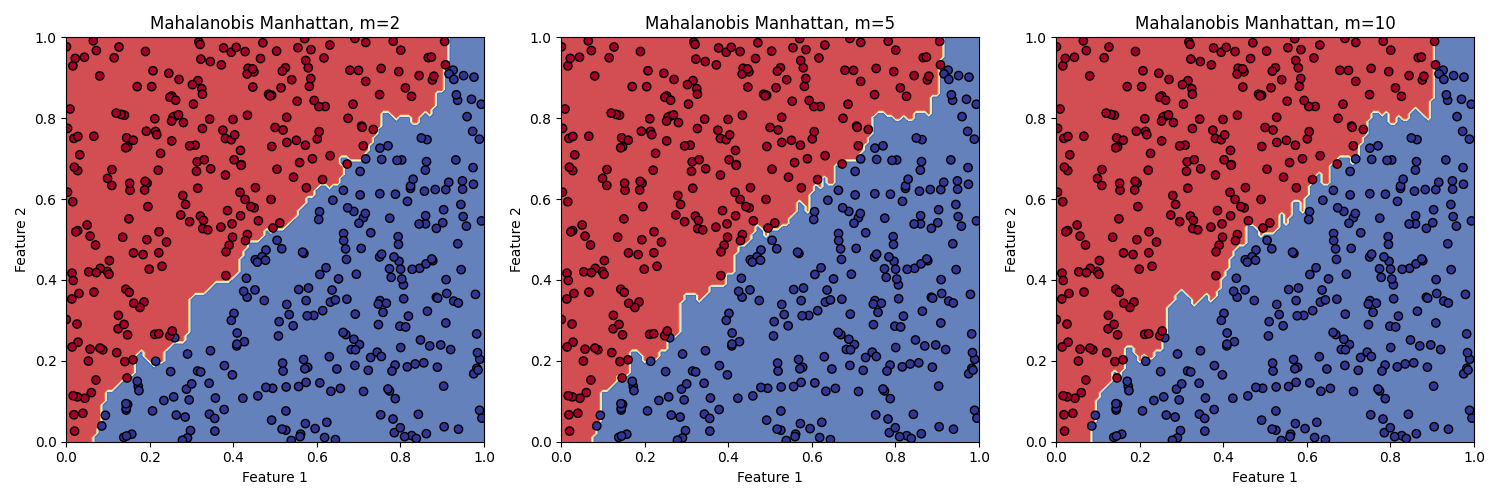} 
        \caption{Mahalanobis Manhattan distance for \(m=2,5,10\)}
    \end{subfigure}
    \centering
    \begin{subfigure}{\textwidth}
        \centering
        \includegraphics[width=\textwidth]{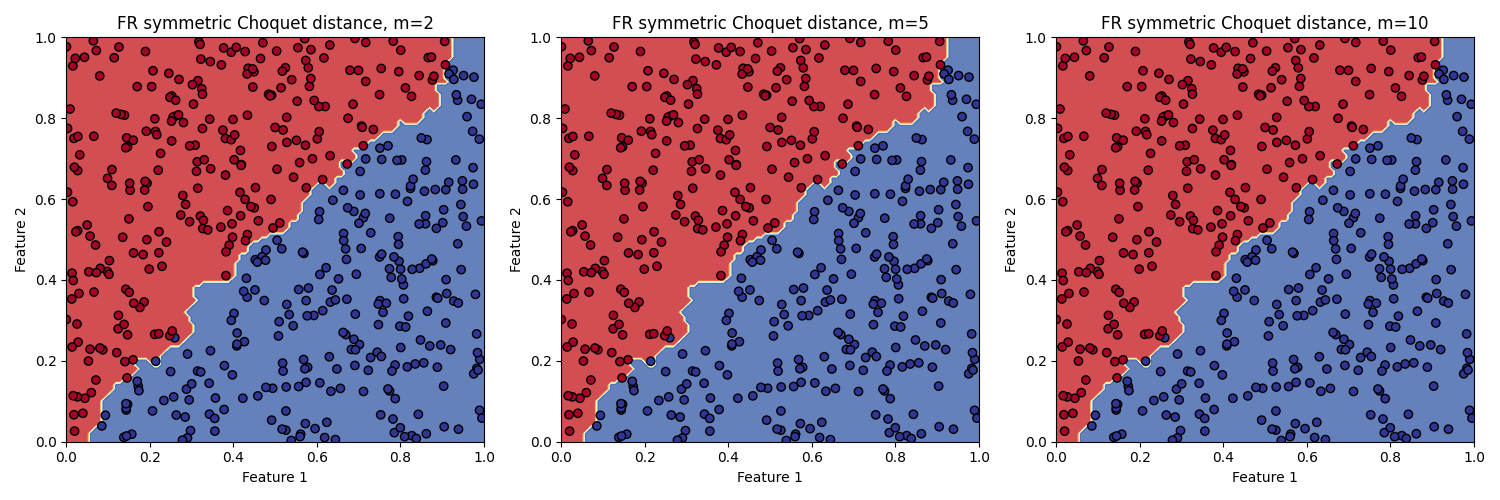} 
        \caption{FR symmetric Choquet distance for \(m=2,5,10\)}
    \end{subfigure}
    \caption{Classification boundaries of the last two distances in a 5-NN setting.}
    \label{fig:visualBoundary2}
\end{figure}

The synthetic dataset is constructed using two informative features, \(a_1: X \to [0,1]\) and \(a_2: X \to [0,1]\), along with \(m\) duplicates \(\{a_3,\dots, a_{m+2}\}\) of \(a_2\). The classification of an instance is defined as:
\[
d(x) =
\begin{cases} 
    1, & \text{if } a_1(x) \geq a_2(x), \\
    0, & \text{otherwise}.
\end{cases}
\]
The training set consists of 500 samples, where the informative features \(a_1\) and \(a_2\) are uniformly distributed in \([0,1]\), i.e., \(a_1,a_2 \sim U[0,1]\). Figure \ref{fig:visualBoundary1} and \ref{fig:visualBoundary2} show the decision boundaries of the different distances for \(m = 2, 5, \text{ and } 10\), representing the number of duplicates of \(a_1\).The optimal decision boundary is the first bisector (\(a_1 = a_2\)). As shown in Figures \ref{fig:visualBoundary1} and \ref{fig:visualBoundary2}, the decision boundary of CFR\textsubscript{.5} remains the closest to this optimal boundary across all values of \(m\). Furthermore, CFR\textsubscript{.5} exhibits perfect stability across different values of \(m\), while the boundaries of MAN and MI deteriorate as \(m\) increases. Although the decision boundary of MAH\textsubscript{1} remains relatively stable, it experiences a slight degradation with increasing \(m\).

\begin{figure}[ht!]
	\centering
	\includegraphics[width=\textwidth]{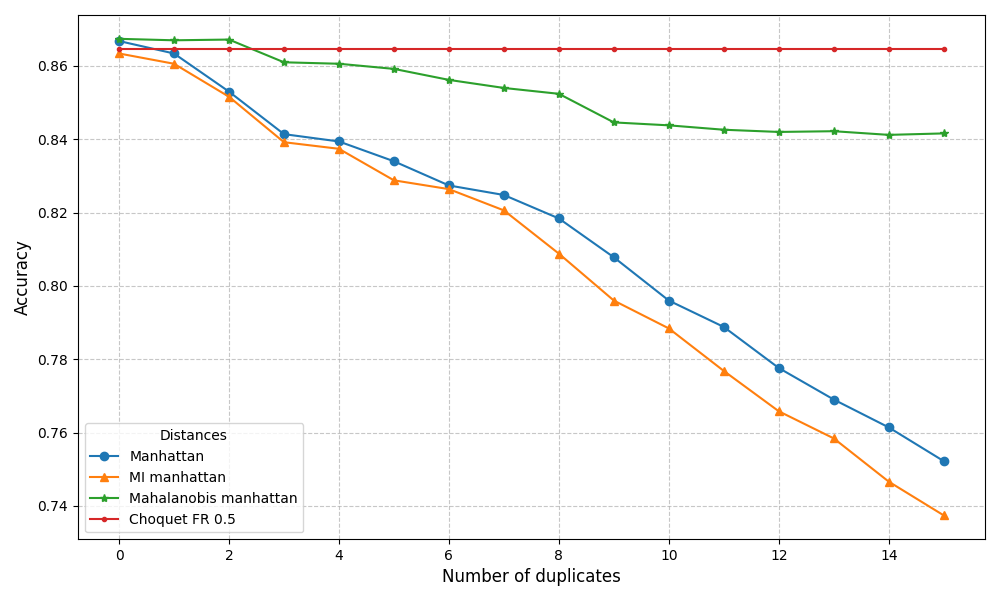}
	\caption{Accuracy in the boundary region for different number of duplicates \(m\)}
	\label{fig:plotAccDuplicates}
\end{figure}

\begin{figure}[ht!]
	\centering
	\includegraphics[width=\textwidth]{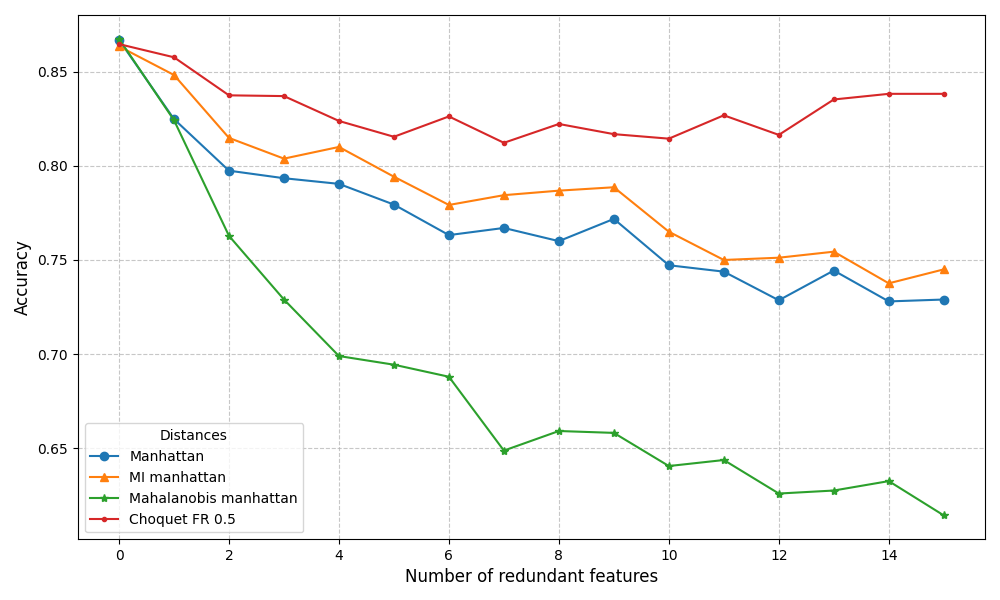}
	\caption{Accuracy in the boundary region for different number of redundant variables}
	\label{fig:strongCorrAttr}
\end{figure}

To fortify this claim, we calculated the accuracy of the proposed methods in the region around the decision boundary by generating a test set of size 5000 near the boundary:
\[
(a_1(x_i), a_2(x_i)) = (y_i + dx_i, y_i + dy_i), \quad y_i \sim U[0,1], \quad dx_i, dy_i \sim U[-0.1,0.1],
\]
for \(i \in \{1,2,\dots, 5000\}\).
The results for \(m\) values ranging from \(0\) to \(15\) are displayed in Figure \ref{fig:plotAccDuplicates}. These results show that CFR\textsubscript{.5} is unaffected against adding duplicates, whereas the performance of Manhattan and MI-weighted Manhattan distances deteriorates quickly. The performance of MAH\textsubscript{1} experiences a slight degradation with increasing \(m\). Figure \ref{fig:strongCorrAttr} illustrates the outcome when strongly correlated attributes are added instead of duplicates. Specifically, each attribute \(a_i(x)\) is defined as \(a_2(x) + \epsilon_{i,x}\), where \(\epsilon_{i,x}\) (\(i \in \{3, \dots, m+2\}\) and \(x \in X\)) are independent and identically distributed (i.i.d.) random variables following a normal distribution with a mean of zero and a standard deviation of \(0.1\). We observe the same overall trend as before, with three notable differences: (1) the MI-weighted Manhattan distance now outperforms the standard Manhattan distance; (2) CFR\textsubscript{.5} now exhibits a slight decline in accuracy before stabilizing towards the end, whereas previously it was completely unaffected by these additional attributes; and (3) the accuracy of MAH\textsubscript{1} deteriorates more rapidly as \(m\) increases.

The first observation can be explained by the fact that \(a_3, \dots, a_{m+2}\) are noisy and, therefore, individually less predictive. As a result, the MI-weighted distance assigns them lower weights. The deterioration of the accuracy for CFR\textsubscript{.5} and MAH\textsubscript{1} is likely due to the introduction of noisy attributes.The limited deterioration in the accuracy of CFR\textsubscript{.5} will be explained in the next subsection.
In conclusion, CFR\textsubscript{.5} remains robust to the inclusion of redundant variables and ultimately outperforms the other methods.


\subsubsection{Discussion}
The performance deterioration of Manhattan and weighted Manhattan distances with the addition of duplicates is evident. Adding duplicates of \(a_1\) increases the weight assigned to \(a_1\) without contributing any new information. In contrast, the Choquet distance remains unaffected because it effectively recognizes that adding duplicates does not enhance the informational content.

Indeed, if we use a measure \(\mu\) that effectively captures the fact that \(\{a_3, \dots, a_{m+2}\}\) are duplicates of \(a_1\) (cf.\ Definition \ref{defn:duplicates}), then by applying Proposition \ref{prop: duplicateChoquet} \(m\) times, we obtain the following expression for the Choquet distance:  
\begin{align*}
	d^\mu(x, y) &= \mu(\{a_1\})\cdot|a_1(x) - a_1(y)| + \mu(\{a_2\})\cdot|a_2(x) - a_2(y)|\\
	&+ \left(\mu(\{a_1,a_2\})-\mu(\{a_1\})-\mu(\{a_2\})\right) \cdot \min(|a_1(x) - a_1(y)|,|a_2(x) - a_2(y)|).
\end{align*}
This formulation is robust against the addition of duplicates, ensuring that redundant features do not disproportionately influence the distance computation. Note that the same reasoning also provides an explanation for the stability observed when strongly correlated attributes are added. Ideally, \(a_2\) and one of its highly correlated features, \(a_i\), are duplicates in the sense of Definition \ref{defn:duplicates}. But, even if they are not exact duplicates, a well-suited measure \(\mu\) would satisfy a weaker form of Definition \ref{defn:duplicates} (\(\gtrapprox\), as \(a_i\) is the noisy attribute): 
\[
\mu(A \cup \{a_2\}) \gtrapprox \mu(A \cup \{a_i\}),
\]
and thus, by the same reasoning as in the proof of Proposition \ref{prop: duplicateChoquet}, the Choquet distance would assign proportionally less weight to the redundant feature \(a_i\).
However, in our experiment we have used CFR\textsubscript{.5} that uses a symmetrized version of the measure defined in Equation \eqref{eq: gamma measure}. But as can be seen directly from Equation \eqref{eq: gamma measure}, if a duplicate \(b\) is added to a set \(B\) the gamma measure remains unchanged:
\begin{align*}
	\gamma_d(B\cup\{b\}) = \sum\limits_{y \in X}\min\limits_{z \notin R_dy}\left(\max\limits_{a\in B\cup\{b\}}|a(z)-a(y)|\right) =\sum\limits_{y \in X}\min\limits_{z \notin R_dy}\left(\max\limits_{a\in B}|a(z)-a(y)|\right)= \gamma_d(B),
\end{align*}
hence having the property of Definition \ref{defn:duplicates}. And if two attributes are duplicates w.r.t.\ \(\mu\) they are also duplicates w.r.t.\ the dual of \(\mu\) (Proposition \ref{prop: duplicateChoquet}), and hence the symmetrized \(\mu^s\).
In conclusion, the Choquet distance provides an elegant approach to account for duplicate features. Furthermore, it effectively handles redundant features, such as highly correlated ones, by appropriately adjusting their effect on the total distance.

\subsection{Experiment: benchmark datasets}
\label{sec: benchmark}
In this subsection, we evaluate the effectiveness of feature subset weighting using Choquet distances by comparing their classification accuracy against traditional feature-weighted distances and several Mahalanobis distance variants on benchmark datasets.
\subsubsection{Experimental Setup}
To assess the performance of the proposed Choquet distances, we perform K-Nearest Neighbors (KNN) classification using Choquet distances, standard weighted distances and several Mahalanobis distances. A summary of the evaluated distances is provided in Table \ref{tab:distances}.
For the implementation of KNN, as well as the \(\chi^2\)-weighted distance and mutual information-weighted distance, we use the scikit-learn library \cite{scikit-learn}. We set \(K = 5\), as this is the default value for KNN in the scikit-learn implementation. Nonetheless, comparable results are observed for other values of \(K\). For the covariance matrix in the Mahalanobis distances, we use the ShrunkCovariance implementation from scikit-learn with default parameters, as alternative parameter choices did not improve its performance.
We will conduct 5-fold cross-validation on 25 datasets (Table \ref{table: datasets}) from the UCI Machine Learning Repository \cite{Dua:2019}, using only numerical features. Balanced accuracy will be employed as the performance metric. 

\begin{table}[h]
    \centering
    \begin{tabular}{ll}
        \toprule
        \textbf{Distance Metric} & \textbf{Description} \\
        \midrule
        MAN  & Standard Manhattan distance \\
        CHI  & \(\chi^2\)-weighted Manhattan distance \\
        MI   & Mutual information-weighted Manhattan distance \\
        MAH  & Mahalanobis distance \\
        MAH\textsubscript{1} & Mahalanobis Manhattan distance (whitened Manhattan) \\
        MAMI & MI-weighted Mahalanobis Manhattan distance \\
        WFR  & Fuzzy rough \(\gamma_d\)-weighted distance \\
        CFR, CFR\textsubscript{.5}, CFR\textsubscript{1} & \(\alpha\)-Choquet distance with \(\alpha \in \{0, 0.5, 1\}\) and \(\mu = \gamma_d\) \\
        \bottomrule
    \end{tabular}
	\caption{Summary of the tested distances.}
    \label{tab:distances}
\end{table}
\begin{table}[h!]
    \centering
    \begin{tabular}{@{}lcccc|lcccc@{}}
        \toprule
        \textbf{Name} & \textbf{\#Feat.} & \textbf{\#Inst.} & \textbf{IR} &  & \textbf{Name}   & \textbf{\#Feat.} & \textbf{\#Inst.} & \textbf{IR} \\
        \midrule
        appendicitis  & 7                & 106              & 4.0         &  & iris            & 4                & 150              & 1.0 (3)     \\
        banknote      & 4                & 1372             & 1.2         &  & new-thyroid     & 5                & 215              & 5.0 (3)     \\
        breasttissue  & 9                & 106              & 1.6 (6)     &  & plrx            & 12               & 182              & 2.5         \\
        caesarian     & 5                & 80               & 1.4         &  & post-op         & 8                & 87               & 2.6         \\
        cmc           & 9                & 1473             & 1.9 (3)     &  & qual-bank       & 6                & 250              & 1.3         \\
        coimbra       & 9                & 116              & 1.2         &  & raisin          & 7                & 900              & 1.0         \\
        column        & 6                & 310              & 2.5 (3)     &  & seeds           & 7                & 210              & 1.0 (3)     \\
        fertility     & 9                & 100              & 7.3         &  & somerville      & 6                & 143              & 1.2         \\
        forest-types  & 9                & 523              & 2.3 (4)     &  & transfusion     & 4                & 748              & 3.2         \\
        glass         & 9                & 214              & 8.4 (6)     &  & userknowledge   & 5                & 403              & 5.4 (5)     \\
        haberman      & 3                & 306              & 2.8         &  & warts           & 8                & 180              & 2.0         \\
        ilpd          & 10               & 579              & 2.5         &  & websitephishing & 9                & 1353             & 6.8 (3)     \\
        wisconsin     & 9                & 683              & 1.9         &  &                 &                  &                  &             \\
        \bottomrule
    \end{tabular}
	
    \caption{\label{table: datasets}Summary of the 25 UCI datasets used, all of which consist of numerical features. (\#Feat. = Number of features, \#Inst. = Number of instances, IR = Imbalance Ratio and number of classes)}
\end{table}
\subsubsection{Results and discussion}
\begin{figure}[!ht]
    \centering
    \begin{minipage}[b]{\textwidth}
        \centering
        \includegraphics[width=0.8\textwidth]{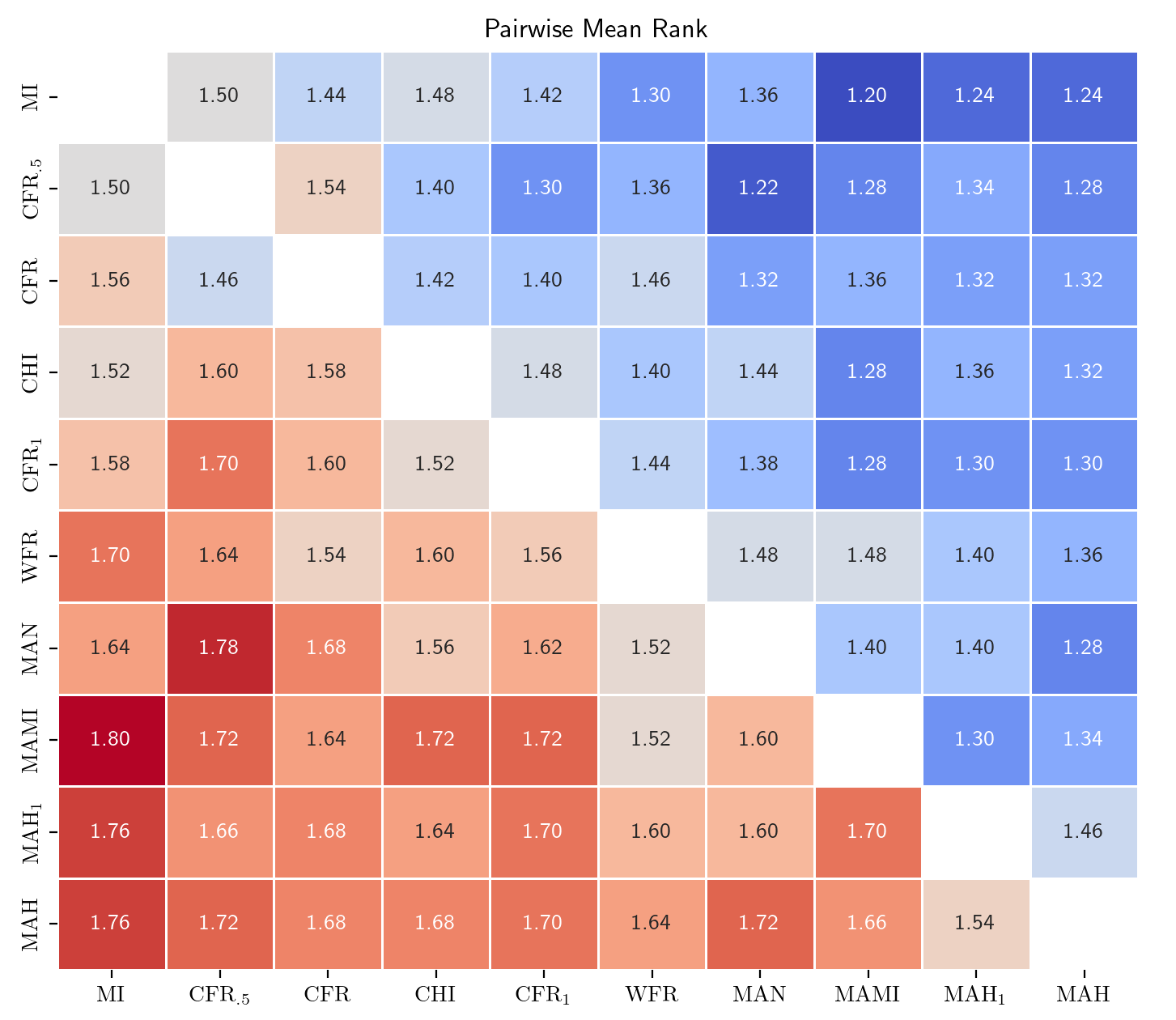}
        \caption{Pairwise mean rank comparison of distance metrics across datasets.}
        \label{fig:pairwise}
    \end{minipage}

    \vskip 1em  
    
    \begin{minipage}[b]{\textwidth}
        \centering
        \includegraphics[width=0.86\textwidth]{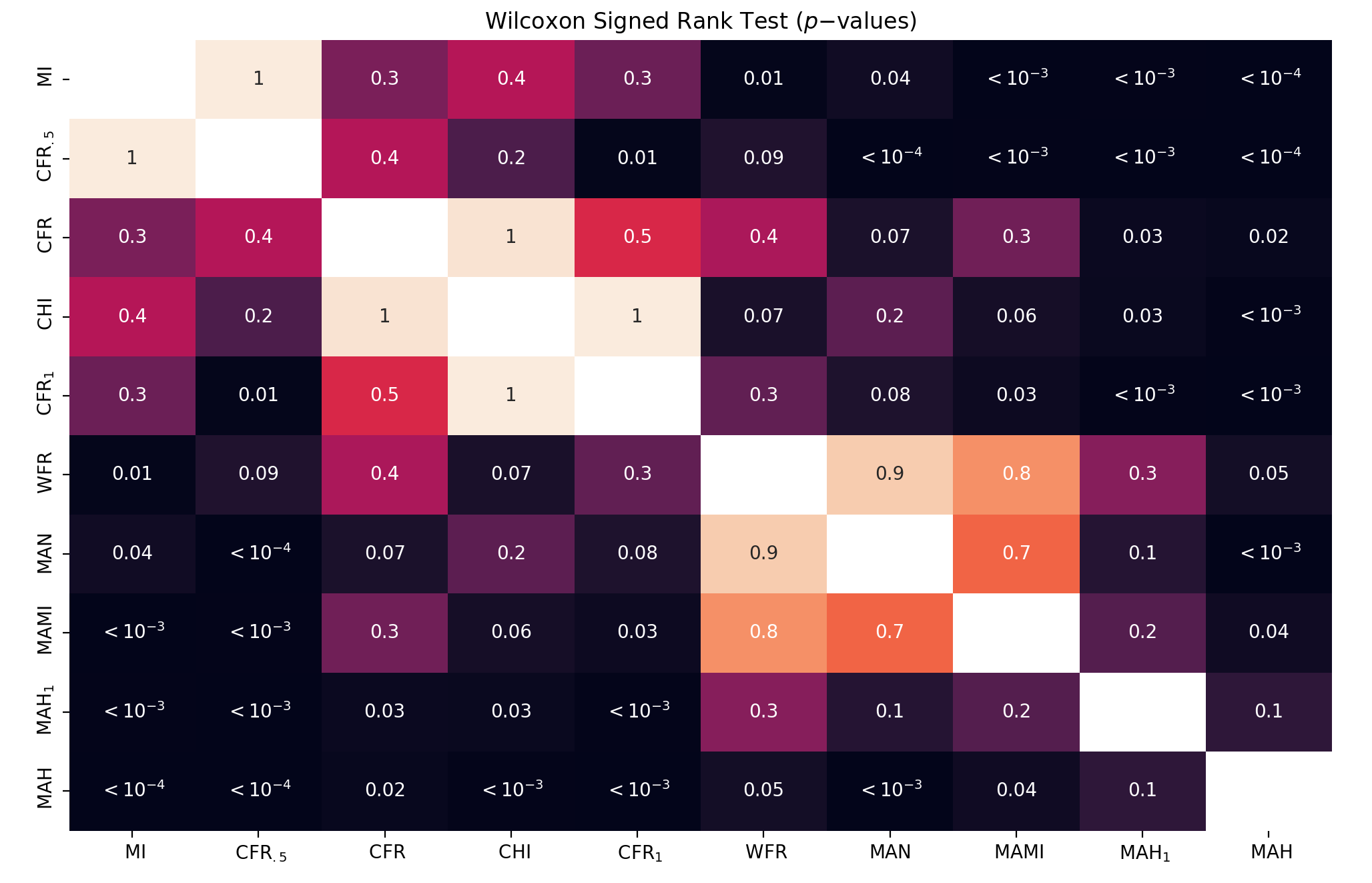}
        \caption{Heatmap of the \(p\)-values from the pairwise two-sided Wilcoxon signed rank test.}
        \label{fig:heatmap}
    \end{minipage}
\end{figure}

The experimental results are summarized in Table \ref{tab:performance_comparison} and Figure \ref{fig:pairwise}. These include the average accuracy, the percentage of datasets where each distance metric outperforms the Manhattan distance, and the pairwise mean ranks. At first glance, we observe that both MI and CFR\textsubscript{.5} achieve the best results, with similar overall performance. However, in terms of outperforming the Manhattan distance, CFR\textsubscript{.5} has an advantage. Furthermore, it is evident that the symmetric Choquet distance outperforms the standard CFR variants. Additionally, we note that the feature subset weighted variants of the fuzzy-rough distance (CFR, CFR\textsubscript{.5}, and CFR\textsubscript{1}) generally outperform the weighted variant (WFR).

\begin{table}[h]
    \centering
	\small
	\begin{tabular}{l r r r r r r r r r r}
		\hline
		\textbf{dataset} & MAN & CHI & MI & MAH\textsubscript{1} & MAH & MAMI & WFR & CFR & CFR\textsubscript{.5} & CFR\textsubscript{1} \\ \hline
		appen. & 0.736 & 0.766 & 0.749 & 0.741 & 0.747 & 0.700 & 0.761 & 0.741 & 0.741 & 0.747 \\ 
		bankn. & 0.998 & 0.984 & 0.993 & 0.996 & 0.997 & \textbf{1.000} & 0.994 & 0.992 & 0.995 & 0.996 \\ 
		breast. & 0.672 & 0.641 & 0.630 & 0.626 & 0.576 & 0.632 & 0.605 & 0.673 & 0.701 & \textbf{0.700} \\
		caesar. & 0.648 & 0.625 & 0.633 & \textbf{0.689} & 0.672 & 0.624 & 0.422 & 0.521 & 0.656 & 0.641 \\ 
		cmc & 0.475 & 0.439 & 0.485 & 0.457 & 0.462 & 0.480 & 0.378 & \textbf{0.491} & 0.480 & 0.485 \\ 
		coimb. & 0.721 & 0.714 & 0.725 & 0.744 & 0.663 & 0.712 & 0.715 & \textbf{0.748} & 0.744 & 0.740 \\ 
		colmn & 0.750 & 0.774 & 0.803 & 0.729 & 0.722 & 0.790 & 0.763 & 0.748 & 0.757 & 0.761 \\ 
		fert. & 0.483 & 0.494 & 0.566 & 0.494 & 0.500 & \textbf{0.622} & 0.522 & 0.494 & 0.522 & 0.533 \\ 
		forest & 0.872 & 0.871 & \textbf{0.878} & 0.858 & 0.863 & 0.851 & 0.861 & 0.858 & 0.851 & 0.850 \\ 
		glass & 0.614 & 0.609 & \textbf{0.686} & 0.590 & 0.591 & 0.627 & 0.609 & 0.609 & 0.667 & 0.632 \\ 
		haber. & 0.575 & 0.555 & 0.539 & 0.576 & 0.576 & 0.538 & \textbf{0.597} & 0.580 & 0.573 & 0.562 \\ 
		ilpd & 0.556 & 0.599 & 0.592 & 0.563 & 0.540 & 0.590 & \textbf{0.607} & 0.587 & 0.605 & 0.589 \\ 
		iris & 0.947 & \textbf{0.953} & 0.947 & 0.907 & 0.913 & 0.913 & \textbf{0.953} & \textbf{0.953} & 0.947 & 0.947 \\ 
		nwthyr. & 0.865 & \textbf{0.925} & \textbf{0.925} & 0.791 & 0.805 & 0.807 & \textbf{0.925} & 0.914 & 0.912 & 0.903 \\ 
		plrx & 0.484 & 0.500 & \textbf{0.533} & 0.506 & 0.497 & 0.449 & 0.490 & 0.503 & 0.492 & 0.487 \\ 
		postop. & 0.461 & 0.468 & \textbf{0.479} & 0.430 & 0.438 & 0.449 & 0.495 & 0.489 & 0.473 & 0.453 \\ 
		qual. & 0.988 & 0.995 & 0.995 & 0.978 & 0.964 & 0.978 & 0.899 & \textbf{1.000} & \textbf{1.000} & 0.996 \\ 
		raisin & 0.853 & 0.849 & 0.842 & \textbf{0.863} & 0.859 & 0.847 & 0.839 & 0.856 & 0.862 & 0.857 \\ 
		seeds & 0.924 & 0.924 & 0.919 & 0.938 & 0.938 & \textbf{0.957} & 0.929 & 0.929 & 0.929 & 0.914 \\ 
		smerv. & 0.510 & \textbf{0.596} & \textbf{0.596} & 0.501 & 0.485 & 0.557 & 0.562 & 0.589 & 0.542 & 0.533 \\ 
		transf. & \textbf{0.627} & 0.598 & 0.611 & 0.593 & 0.612 & 0.602 & 0.610 & 0.618 & 0.616 & 0.617 \\ 
		usr. & 0.694 & \textbf{0.832} & 0.802 & 0.680 & 0.631 & 0.783 & 0.763 & 0.756 & 0.794 & 0.793 \\ 
		warts & 0.815 & 0.823 & 0.802 & 0.781 & 0.760 & 0.786 & 0.793 & \textbf{0.836} & 0.823 & 0.819 \\ 
		webphis. & 0.748 & 0.802 & 0.784 & 0.774 & 0.731 & 0.782 & 0.751 & 0.630 & 0.858 & \textbf{0.861} \\ 
		wisc. & 0.964 & 0.964 & 0.964 & 0.931 & 0.943 & 0.936 & 0.954 & 0.962 & 0.960 & 0.954 \\  \hline \hline
		average & 0.719 & 0.732& 0.739& 0.710 & 0.699 & 0.721& 0.712 & 0.723 & \textbf{0.740} & 0.733 \\
        \(\%\geq\) man & - & 0.56 & 0.68&0.40 & 0.28& 0.40& 0.52 & 0.68 & \textbf{0.80} & 0.64 \\
	\end{tabular}
	\caption{Mean balanced accuracy results from performing 5-fold cross-validation.}
    \label{tab:performance_comparison}
\end{table}
To determine whether any of these methods consistently and significantly outperforms another, we conduct a two-sided Wilcoxon signed-rank test. The results of this analysis are presented in Figure \ref{fig:heatmap}.
First, we observe that CFR\textsubscript{.5} significantly outperforms the Manhattan distance (MAN) with near perfect significance (\(p < 10^{-4}\)), whereas MI only shows moderate significance in its outperformance (\(p = 0.04\)). Furthermore, CFR\textsubscript{.5} marginally outperforms (\(p = 0.09\)) its weighted variant (WFR). Additionally, MI significantly outperforms WFR (\(p = 0.01\)).The Mahalanobis distances performed poorly, with MAH consistently underperforming compared to the Manhattan distance (\(p<10^{-3}\)). However, MAMI showed a significant improvement over the Mahalanobis distance (\(p = 0.04\)).
\subsection{Conclusion}
Although CFR\textsubscript{.5} does not consistently outperform classical weighted approaches, its superior performance over the Manhattan distance establishes it as a strong competitor. Given the substantial improvement from WFR to CFR\textsubscript{.5}, along with the consistent superiority of MI over WFR, we may cautiously infer that feature subset weighting offers performance advantages over simple feature weighting. While weighting the Mahalanobis distance by mutual information and using its Manhattan variant showed significant improvement, the approach still performed poorly compared to the other methods. In particular, CFR\textsubscript{.5} outperformed the Mahalanobis distances significantly.

When it comes to handling duplicates and strongly correlated features, the Choquet distances outperformed both the weighted distances and the Mahalanobis distances.

\FloatBarrier
\section{Conclusion and future work}
\label{sec: conclusion}
Feature subset weighting using the Choquet integral provides an interpretable approach for incorporating higher-order correlation effects between conditional attributes and the decision attribute. Our analysis demonstrates that feature subset weighting methods, particularly the symmetric Choquet distance, have the potential to outperform traditional feature weighting techniques. Specifically, we observed that extending the weighted distance based on the fuzzy rough dependency measure to a Choquet distance significantly improves its performance. This highlights the strength of the Choquet distance in capturing intricate relationships between features. 

Although the Mahalanobis distance is a flexible and powerful distance when combined with metric learning, its performance diminishes when used with predefined parameters, resulting in poor results, as observed in our experiments. In contrast, the Choquet distance retains its flexibility and achieves superior performance without the need for explicit parameter learning, all while preserving the original features and enhancing interpretability. Given its ability to outperform the Mahalanobis distance without weight optimization, incorporating metric learning with the Choquet distance could further boost its performance and adaptability.

Furthermore, a key advantage of the Choquet distance is its effective handling of feature redundancy. As demonstrated, it outperforms both weighted distances and the Mahalanobis distance in managing duplicates and strongly correlated features. By aggregating feature contributions in a non-additive manner, it inherently mitigates the influence of highly correlated or duplicate features, eliminating the need for explicit preprocessing steps.

While these findings underscore the potential of subset weighting, further research is needed to enhance its effectiveness and explore its full range of applications. First and foremost, reducing time complexity remains a top priority. One promising approach is the utilization of \(k\)-additive measures. Additionally, performance could be further enhanced by extending mutual information-based feature weighting methods to the weighting of feature subsets. Another intriguing direction for future research lies in leveraging \(\lambda\)-fuzzy measures or, more broadly, distorted probability measures. Finally, a worthwhile investigation would be to assess whether the Choquet distance, particularly when restricted to the class of \(2\)-additive measures, can serve as an innovative framework for metric learning, offering a viable alternative to the Mahalanobis distance.

%
%
%

\end{document}